\pdfoutput=1

\documentclass[11pt]{article}

\usepackage[preprint]{acl}

\usepackage{times}
\usepackage{latexsym}

\usepackage[T1]{fontenc}

\usepackage[utf8]{inputenc}

\usepackage{microtype}

\usepackage{inconsolata}

\usepackage{graphicx}

\usepackage{url}

\usepackage{booktabs}       
\usepackage{amsfonts}       
\usepackage{nicefrac}       
\usepackage{microtype}      

\usepackage{multirow}
\usepackage{graphicx}
\usepackage{float}
\usepackage{wrapfig}
\usepackage{caption}
\usepackage{subcaption}
\usepackage{enumitem}
\usepackage{array}

\usepackage{amssymb}
\usepackage{amsmath}
\usepackage{bbm}
\usepackage{bbding}
\usepackage{appendix}

\usepackage{hyperref} 

\usepackage{xcolor}

\usepackage[capitalize]{cleveref}
\crefname{section}{Sec.}{Secs.}
\Crefname{section}{Section}{Sections}
\Crefname{table}{Table}{Tables}
\crefname{table}{Tab.}{Tabs.}

\def\eg{\emph{e.g.}}
\def\ie{\emph{i.e.}}

\def\mmethod{StructTuning}
\newcommand{\medit}[1]{#1}

\usepackage{amsmath,amsfonts,bm}

\def\eqref#1{equation~\ref{#1}}

\def\1{\bm{1}}

\def\vs{{\bm{s}}}

\def\vx{{\bm{x}}}

\def\vz{{\bm{z}}}

\DeclareMathAlphabet{\mathsfit}{\encodingdefault}{\sfdefault}{m}{sl}
\SetMathAlphabet{\mathsfit}{bold}{\encodingdefault}{\sfdefault}{bx}{n}

\title{Structure-aware Domain Knowledge Injection for Large Language Models}

\author{
Kai Liu$^{1,2}$\footnotemark[1],\; Ze Chen$^{2}$\footnotemark[1],\;  Zhihang Fu$^{2}$\footnotemark[2],\; Wei Zhang$^{1}$,\; Rongxin Jiang$^{1}$,\  \\
\textbf{Fan Zhou}$^{1}$\footnotemark[2],\; \;\textbf{Yaowu Chen}$^{1}$,\; \textbf{Yue Wu}$^{2}$,\;  \textbf{Jieping Ye}$^{2}$ \vspace{0.7em} \\
$^{1}$Zhejiang University, $\;$ $^{2}$Alibaba Cloud \\
}

\begin{document}

\maketitle

\renewcommand{\thefootnote}{\fnsymbol{footnote}}
\footnotetext[1]{Equal contribution. Work done during Kai Liu's research internship at Alibaba Cloud. Email: kail@zju.edu.cn.}
\footnotetext[2]{Corresponding authors. Email: zhihang.fzh@alibaba-inc.com, 0006236@zju.edu.cn.}
\renewcommand{\thefootnote}{\arabic{footnote}}

\begin{abstract}
This paper introduces a pioneering methodology, termed \textit{\mmethod}, to efficiently transform foundation Large Language Models (LLMs) into domain specialists.
\medit{It significantly reduces the training corpus needs to a mere \textbf{5\%} while achieving an impressive \textbf{100\%} of traditional knowledge injection performance.}
Motivated by structured human education, we propose a novel two-stage strategy for knowledge injection and alignment: \textit{Structure-aware Continual Pre-Training} (SCPT) and \textit{Structure-aware Supervised Fine-Tuning} (SSFT). 
In the SCPT phase, we automatically extract the domain knowledge taxonomy and reorganize the training corpora, enabling LLMs to effectively link textual segments to targeted knowledge points within the taxonomy. 
In the SSFT phase, we explicitly prompt models to elucidate the underlying knowledge structure in their outputs, leveraging the structured domain insight to address practical problems.
Our ultimate method was extensively evaluated across model architectures and scales on LongBench and MMedBench datasets, \medit{demonstrating superior performance against other knowledge injection methods}. 
We also explored our method's scalability across different training corpus sizes, laying the foundation to enhance domain-specific LLMs with better data utilization.
Code is available at:
\url{https://github.com/alibaba/struxgpt}.
\end{abstract}

\section{Introduction}
\label{sec:intro}

\begin{figure*}[t]
\begin{center}
\centerline{\includegraphics[width=0.85\linewidth]{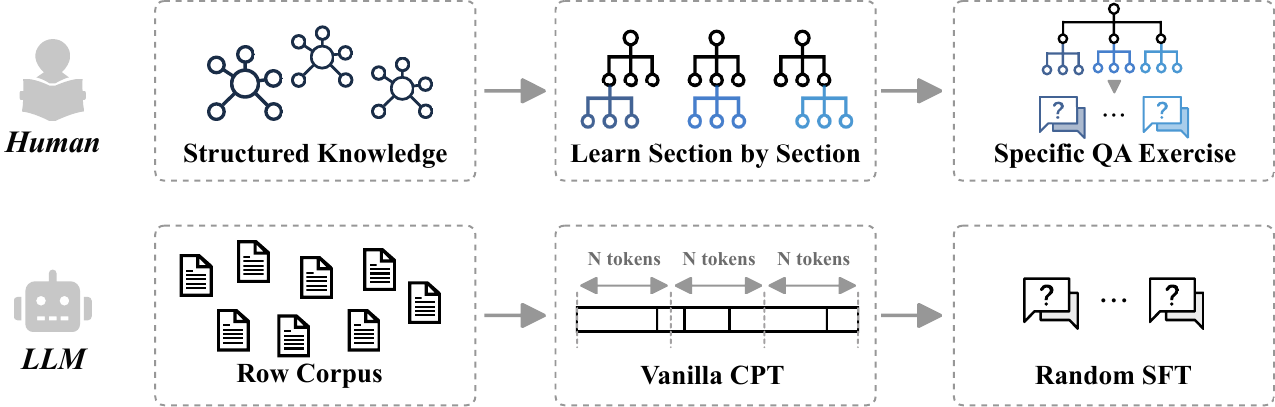}}
\caption{\textbf{Discrepancy between human education and vanilla LLM adaptation}. Human students learn structured knowledge through textbooks section by section, with particular exercises on related knowledge points. Traditional LLM adaptation continually pre-trains on data chunks from randomly concatenated text segments, with aimless supervised fine-tuning for conversation alignment. The inherent knowledge structure is ignored.}
\label{fig:intro}
\end{center}
\vskip -0.3in
\end{figure*}

Large language models (LLMs) have recently seen extensive deployment across various applications~\citep{vaswani2017attention,achiam2023gpt,jiang2023mistral,bi2024deepseek}. 
When adapting foundational models (\eg, Llama series~\citep{touvron2023llama,touvron2023llama2,dubey2024llama}) to specialized AI assistants in distinct domains, developers usually employ two techniques to enhance LLMs' proficiency: retrieval-augmented generation (RAG)~\citep{lewis2020rag} and domain knowledge injection~\citep{gururangan2020inject}. 
While RAG effectively utilizes an external knowledge base to augment information, the retrieval process's inherent noise poses challenges to generating reliable responses, especially in scenarios requiring logical reasoning where there is a semantic gap between the user's query and the knowledge base.\citep{zhang2023siren,chen2023understanding}. 
Thus, another avenue tries to inject new knowledge to LLMs via training techniques~\citep{gu2021domain,hu2021lora,mecklenburg2024injecting}.

Continual pre-training~\citep{sun2020ernie,ibrahim2024simple} is widely used for injecting domain-specific knowledge~\citep{cui2023chatlaw,wang2023huatuo,qiu2024mmedlm}. 
However, it often requires training on billions of internet tokens to capture fragmented knowledge, rather than leveraging a few structured textbooks~\citep{jin2020disease}. 
For instance, MMedLM~\citep{qiu2024mmedlm} uses 25.5B tokens for medical modeling, while DeepSeek-Coder~\citep{guo2024deepseekcoder} processes 2T tokens for coding adaptation. 
The limited ability to learn effectively from textbooks was attributed to insufficient data diversity~\citep{zhu2023physics}, which however violates the observation during the human education process in \cref{fig:intro}: students gain knowledge by sequentially studying from textbooks, reviewing knowledge points and structures, and applying this knowledge through proper exercises.
Here, all the new data to learn are textbooks (structured content) and exercising examples (question-answering pairs), and students just adopt their world knowledge to memorize, understand, and apply the knowledge to become domain experts~\citep{krathwohl2002revision,yu2023kola}.

As educating human students, we propose to inject structured domain knowledge into LLMs via two steps: \textit{Structure-aware Continual Pre-Training} (SCPT) and \textit{Structure-aware Supervised Fine-Tuning} (SSFT). 

In the SCPT stage, we argue that high-quality textbook data (as well as regular corpora from the Internet) can adequately infuse domain knowledge~\citep{gunasekar2023textbooks}, where the organization of training corpora is crucial. 
In conventional paradigms (\cref{fig:intro}), text corpora are simply concatenated and divided into chunks of 2048~\citep{qiu2024mmedlm} or 4096~\citep{guo2024deepseekcoder} tokens, while the inherent semantic structure (\eg, catalogs of textbooks) is discarded. 
Instead, we view each chunk as a knowledge point and automatically extract domain knowledge taxonomy from the whole corpus. 
Subsequently, LLMs are trained to predict textual content (corresponding to a knowledge point) \textit{under the condition of} the knowledge path within the domain structure, linking individual training chunks with the entire knowledge architecture. 
Finally, models have to memorize the entire structure to review the whole domain knowledge system.

In the SSFT stage, the goal shifts from knowledge injection to enabling LLMs to recall and utilize their acquired knowledge to tackle real-world challenges. 
We explicitly elicit knowledge paths in LLMs' responses, as a beacon for models to targeted information retrieval or logical reasoning for reliable responses. 
To this end, we derive a scalable strategy to generate question-answer pairs as practice exercises by open-sourced LLMs or API, such as LLaMA3~\citep{dubey2024llama} and GPT4~\citep{achiam2023gpt}.
In the scenarios with existing QA pairs like MMedBench~\citep{qiu2024mmedlm}, we retrieve the related knowledge structure and content, instructing LLaMA3 to provide explanations from questions to answers based on the knowledge paths. 
For datasets lacking specific QA samples like LongBench~\citep{bai2023longbench}, we randomly select knowledge paths from the domain taxonomy and prompt LLaMA3 to craft question-answer-explanation triplets for training exercises.

Our ultimate approach \textit{\mmethod}~has been extensively evaluated across different model architectures and sizes. 
In particular, we first examine their capability to recall the knowledge injected through open-ended QA on the LongBench~\citep{bai2023longbench} dataset, then assess the application of injected knowledge to address real-world issues through multiple-choice QA on MMedBench~\citep{qiu2024mmedlm}. 
Both evaluations underscore the superiority of \mmethod, surpassing other SOTA domain knowledge injection methods~\citep{cheng2023adapting, zhang2024raft}. 
Remarkably, we achieve a \textbf{50\%} improvement in knowledge injection compared to SOTA MMedLM2 in the medical domain, using only \textbf{0.3\%} of the training data requirement. 
Furthermore, \mmethod~exhibits good scalability, achieving comparable performance with only \textbf{5\%} of the training data.
These findings reveal our superiority in enhancing domain-specific AI assistants with more efficient data utilization.

\begin{figure*}[!t]
\begin{center}
\centerline{\includegraphics[width=0.95\linewidth]{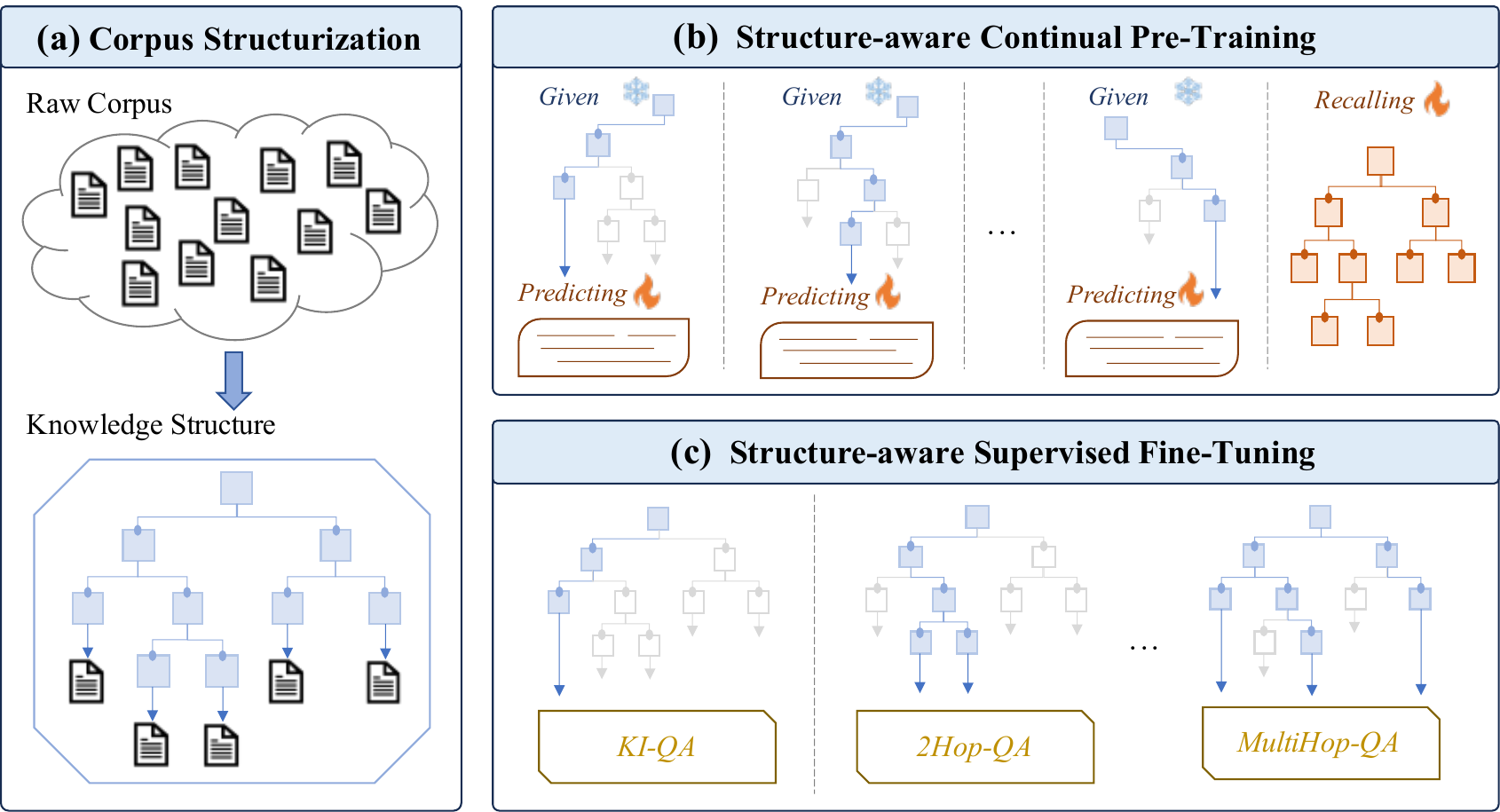}}
\caption{\textbf{Framework for structure-aware knowledge injection}. We extract the inherent knowledge structure in the training corpus, and associate training chunks to corresponding knowledge points. Models are continually pre-trained on data chunks in the condition of the knowledge structure, and fine-tuned with supervised QA samples to elicit their learned knowledge to solve knowledge-intensive (KI) and 2- or multi-hop questions in the real world.}
\label{fig:fmwk}
\end{center}
\vskip -0.3in
\end{figure*}

Our contribution is summarized as follows:

\begin{itemize}[partopsep=2pt,itemsep=3pt,topsep=0pt,parsep=0pt,leftmargin=20pt]
    \item We proposed a novel two-stage training strategy, SCPT and SSFT, to inject domain knowledge into LLMs by preserving and utilizing the inherent structure of the training corpus.
    \item We developed a scalable data construction framework to generate structure-aware training samples from original corpora to facilitate the SCPT and SSFT stages.
    \item We conducted extensive investigations on our \mmethod~strategy on various data and model settings, and comprehensively illustrate our superiority in knowledge injection.
\end{itemize}

\section{Related Works}
\label{sec:rel}

Here we briefly discuss the closely related works. A detailed discussion can be found in \cref{app:sec:related_work}.

\indent \textbf{Domain Adaptation for LLMs}.
To address the domain adaptation problem, a pre-trained model will be continually pre-trained (CPT) with domain-specific content~\citep{sun2020ernie,xu2023kilm}, and fine-tuned with supervised instruction-response pairs (SFT) to keep advancing interactive capabilities~\citep{mecklenburg2024injecting,qiu2024mmedlm}. 
This paradigm is validated efective in dynamic fields like 
medicine~\citep{wang2023huatuo,qiu2024mmedlm} 
and coding~\citep{roziere2023codellama,guo2024deepseekcoder}. 
Our study builds upon this CPT-SFT framework, innovating with SCPT-SSFT strategies to efficiently and effectively infuse domain knowledge with the inherent structure hierarchy.

\textbf{Structure-aware Knowledge Aggregation}.
In conventional paradigms, researchers extract entity-relation-entity triplets from texts to construct knowledge graphs~\citep{pan2024unifying}, to enhance LLMs's factual knowledge and logical reasoning\citep{zhang2022dkplm,wen2023mindmap}. 
Here, each node corresponds to either a specific entity or an abstract concept, lacking the capability to present an informative and self-contained \textit{knowledge point}.
This paper extends to structure-aware knowledge aggregation on existing training corpora, injecting the whole domain knowledge structure into LLMs' by linking training samples to corresponding knowledge points and reasoning paths.

\textbf{Data Augmentation and Synthesis}.
Traditional methods aim to artificially expand the training dataset size~\citep{xu2023wizardlm,mukherjee2023orca} or generate entirely new samples to adapt LLMs to specific tasks~\citep{tang2024mathscale}. 
Yet, they often overlook the structured nature of domain knowledge, while the aimlessly generated samples may also lack diversity~\citep{ovadia2023fine,mecklenburg2024injecting} and cannot cover the domain knowledge points~\citep{mecklenburg2024injecting,tang2024mathscale}.
By contrast, our SSFT design is an innovative departure to address the challenge of retaining and utilizing the structured knowledge inherent in domain-specific content.
\section{Methodology}
\label{sec:method}

\begin{figure*}[htb]
\begin{center}
\centerline{\includegraphics[width=0.9\linewidth]{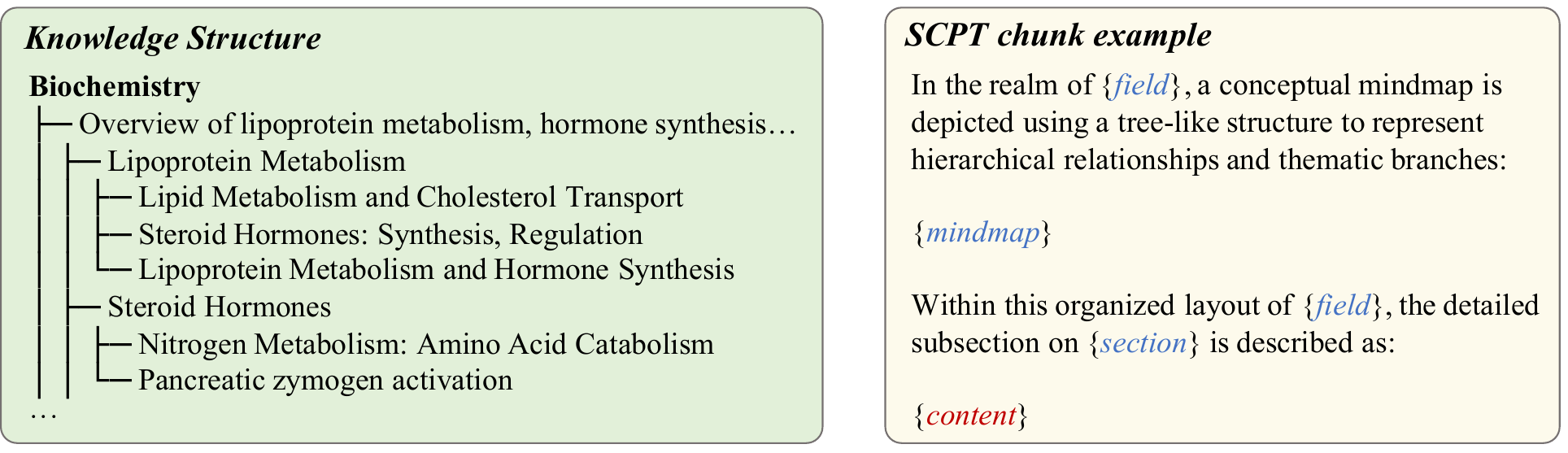}}
\caption{\textbf{Left}: extracted knowledge structure. \textbf{Right}: template to bridge mindmap structure and textual contents.}
\label{fig:struct_isnt}
\end{center}
\vskip -0.3in
\end{figure*}

\cref{fig:fmwk} depicts our \mmethod~methodology to inject domain knowledge into pre-trained LLMs using the inherent knowledge structure.
With curated domain corpora (typically a few textbooks), we first extract the knowledge structure, and associate text chunks to corresponding knowledge paths and points (\cref{sec:method_struct}).
Then, we design a two-stage training strategy to inject the highly structured domain knowledge into language models by mimicking the human education process, comprising the SCPT (\cref{sec:method_scpt}) and SSFT (\cref{sec:method_ssft}) techniques.

\subsection{Knowledge Structure Extraction}
\label{sec:method_struct}

For web-crawled corpus, previous data pre-processing focuses on quality assessment for individual documents~\citep{bi2024deepseek}, while the meta-info of knowledge structures (\eg, the table content for a textbook) is usually neglected or filtered out, and all we have are those sequentially arranged text segments (\eg, page-by-page-chunked content).
As shown in \cref{fig:fmwk} (a), we aim to extract (or, recover) the knowledge structure from the raw corpus for subsequent domain knowledge injection.

First, we use spaCy\footnote{https://github.com/explosion/spaCy} to split the content from a textbook at the paragraph-level, and merge the sentences to form training chunks within a maximum size (\eg, 2048 tokens~\citep{qiu2024mmedlm}).
After that, we prompt the advanced Llama3-70B~\citep{dubey2024llama} model to summarize the title for each chunk, where the textual content with the abstractive title jointly contributes to a ``knowledge point''.

Then, we aggregate knowledge points and extract the inherent structure hierarchy by leveraging advanced language models.
Inspired by \citet{liu2024enhancing}, we take the title list to instruct a specifically developed 7B model to identify the inherent knowledge structure (as exemplified in \cref{fig:struct_isnt}) within the text chunks, and \cref{app:fig:struct_extract} showcases how to deal with non-textbook data.
The whole process does not involve human annotation, which reduces the cost and makes our method scalable for larger domain training corpora.

In particular, \cref{app:sec:struxgpt} and \cref{app:sec:train_cost_cmp} verify that our specialized 7B model can identify sufficiently precise knowledge structure for effective and efficient domain adaptation, as more powerful LLMs like LLaMA3-70B~\citep{dubey2024llama} and GPT-3.5~\citep{brown2020language} cannot bring significant enhancement while largely increase the inference costs.

\subsection{Structure-aware Continual Pre-Training}
\label{sec:method_scpt}

\begin{figure*}[t]
\begin{center}
\centerline{\includegraphics[width=1.0\linewidth]{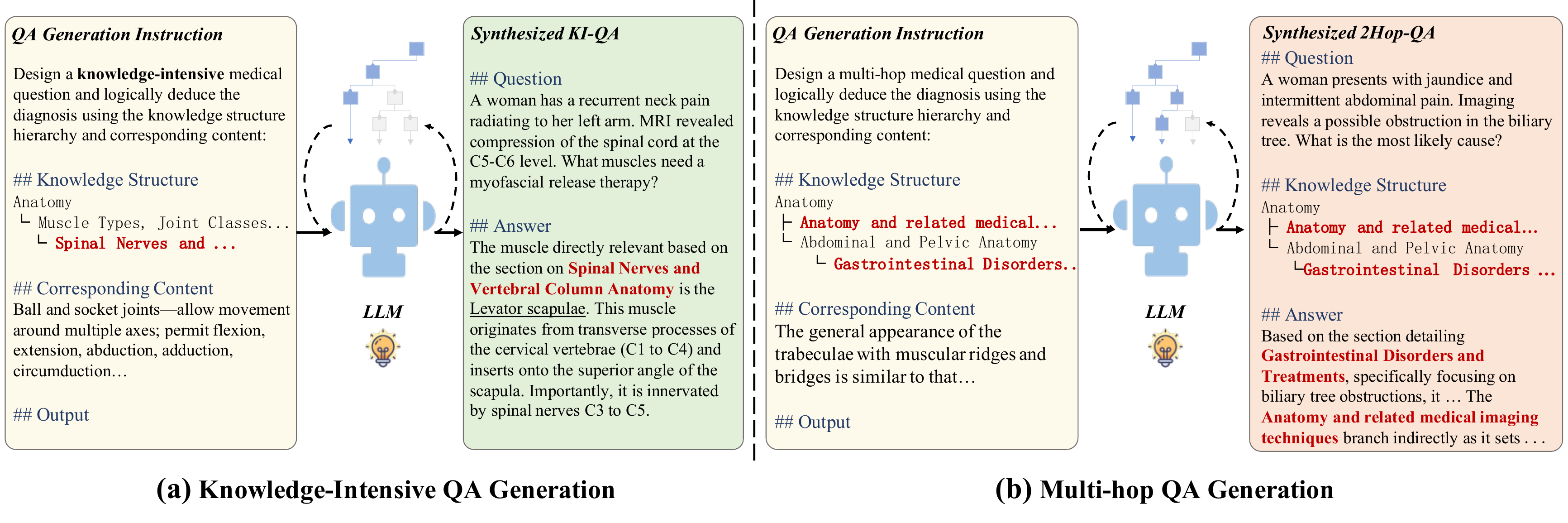}}
\vskip -0.1in
\caption{\textbf{QA samples synthesized for SSFT}. We instruct Llama3-70B to generate (a) knowledge-intensive and (b) multi-hop questions and derive the diagnosis answers with explicit reasoning.}
\label{fig:ssft_sample}
\end{center}
\vskip -0.3in
\end{figure*}

In conventional knowledge injection, training corpora are randomly concatenated and chunked into text segments without distinguishing the original content, making models only able to absorb domain knowledge emerging in data diversity~\citep{ovadia2023fine,mecklenburg2024injecting,qiu2024mmedlm}. 
In this section, we present another solution to inject knowledge from limited text corpora by leveraging the highly abstractive and exhaustive domain knowledge structures for continual pre-training.

We first transform the knowledge structure into natural languages using the same mindmap template~\cite{wen2023mindmap} in \cref{fig:struct_isnt} (left), and prepend it to each training chunk, forcing LLMs to memorize the textual content (knowledge points) in the condition of the associated knowledge path in the structure hierarchy.
20 diversified templates (see \cref{fig:mindmap_templates}) are collected from GPT-4~\citep{achiam2023gpt} to bridge mindmap structures and training chunks, one of which is displayed in \cref{fig:struct_isnt} (right).  
The prepended mindmap, as well as the template, does not produce auto-regressive loss.
Losses are only calculated in the \textit{content} part.
Formally, we turn the original language modeling in vanilla CPT to conditioned modeling~\citep{keskar2019ctrl} in our SCPT stage:

\begin{equation}
\begin{aligned}
  p(\vx^k) = \prod_{i=1}^{n}{p(x^k_i|x^k_{<i})}  \quad \Longrightarrow \quad \\ 
  p(\vx^k|\vs^k) = \prod_{i=1}^{n}{p(x^k_i|x^k_{<i},\vs^k)}
  \label{eq:cond_model}
\end{aligned}
\end{equation}

\noindent where $p(\vx^k)$ models the probability distribution for the $k$-th chunk $\vx^k = (x^k_1, \cdots, x^k_n)$ via the chain rule of probability~\citep{bengio2000neural} on each token $x^k_i$, and $\vs^k$ denotes the associated knowledge mindmap. 
\cref{app:sec:abl_mmed_scpt} extensively investigates the effectiveness of our SCPT strategy.

As \cref{fig:fmwk} (b) shows, after traversing the $m$ knowledge points in extracted structures, models are asked to recall the whole knowledge hierarchy, \ie, to model the composed probability distribution:

\begin{equation}
\begin{aligned}
  p(\bar{\vs}) = \prod_{k=1}^{m}{p(\vs^k)}
  \label{eq:cond_mindmap}
\end{aligned}
\end{equation}

In SCPT, we mimic the human education process to inject knowledge into LLMs in a section-by-section manner, and replay the entire knowledge structure for the models to review and summarize the learned domain knowledge.
These two steps iteratively alternate throughout training epochs.

\subsection{Structure-aware Supervised Fine-Tuning}
\label{sec:method_ssft}

Traditional supervised fine-tuning aims to align the (continually) pre-trained models to interactive ChatBots through question-answering exercises~\citep{cui2023chatlaw,qiu2024mmedlm}.
Previous studies focus on enlarging the quantity and enhancing the diversity of training syntheses~\citep{xu2023wizardlm,mukherjee2023orca,liu2024best} but neglect the highly structured domain knowledge.
In contrast, our structure-aware supervised fine-tuning (SSFT) technique aims to elicit models' structured knowledge learned during SCPT, adapting LLMs to interactive and reliable domain experts.

\cref{fig:fmwk} (c) illustrates SSFT samples synthesis guided by domain knowledge structures.
First, we use the random walk algorithm to create knowledge paths with $1$ to $l$ branches in the original mindmap (the illustration of knowledge paths and branches is displayed in \cref{fig:knowledge_path_example}). 
For paths linking to a single knowledge point, we use the corresponding text content to prompt Llama3-70B~\citep{dubey2024llama} to generate knowledge-intensive QA pairs.
For paths with two or more branches, we prompt Llama3-70B with the knowledge path and textual contents to synthesize 2- or multi-hop QA samples, which require specific reasoning along the knowledge structure to derive from questions to answers.
\cref{fig:ssft_sample} presents several examples.

For every synthesized QA sample ($\vz$), we will prepend the relevant mindmap hierarchy to the answer, and add a CoT prompt in the question to construct another type of QA data ($\vz^\prime$) for SFT alignment.
This design explicitly elicits the learned knowledge in models' responses, teaching them how to apply the structured knowledge to address real-world problems.
We use the two types of QA samples for training, as recommended by \citet{qiu2024mmedlm}.
During testing, we use the vanilla question as input to efficiently gather models' answers to calculate accuracy, and take the CoT prompt to probe to what extent LLMs can memorize and leverage the injected knowledge to answer the questions.

Integrating with SCPT and SSFT, our \mmethod~approach translates into remarkable efficacy and efficiency in domain knowledge injection, as comprehensively evaluated in subsequent sections.

\section{Experiments}
\label{sec:exp}

\begin{table*}[htb]
  \centering
  \caption{\medit{\textbf{Recall of Closed-Book QA (CBQA) on LongBench}}~\citep{bai2023longbench}. The \textbf{best} and \underline{secondary} results are marked in \textbf{bold} and \underline{underlines}, respectively. The base model is Llama2-7B.}
  \label{tab:exp_long_r}
  \vskip -0.1in
  \renewcommand\arraystretch{1.1}
  \small
  \begin{tabular}{c|c|cccccccc}
    \toprule
    \multirow{2}[1]{*}{\textbf{Task}} & \multirow{2}[1]{*}{\textbf{Adaptation}} & \multicolumn{3}{c}{\textbf{SingleDoc-QA}} & \multicolumn{4}{c}{\textbf{MultiDoc-QA}} & \multirow{2}[1]{*}{\textbf{Average}} \\
    \cmidrule(lr){3-5} \cmidrule(lr){6-9}
    & & Qasper & MFQA & MFQAzh & HpQA & 2Wiki & Musiq & Duzh   \\
    \midrule
    \multirow{3}[2]{*}{CBQA}
    &  CPT+SFT  & \underline{20.7} & 35.3 & \underline{20.6} & 29.9 & 32.1 & 18.9 & 12.0 & 24.2 \\
    &  SCPT+SFT    & 18.8 & \underline{42.5} & 17.7 & \underline{35.7} & \underline{36.4} & \underline{20.5} & \underline{15.3} & \underline{26.7} \\
    & SCPT+SSFT & \textbf{30.5} & \textbf{44.6} & \textbf{24.3} & \textbf{40.8} & \textbf{42.0} & \textbf{21.8} & \textbf{16.8} & \textbf{31.5} \\
    \bottomrule
  \end{tabular} 
\vskip -0.1in
\end{table*}

We design extensive evaluations of our \mmethod~through several experiments on two benchmarks.
First, we investigate the free-form question-answering task on the LongBench~\citep{bai2023longbench} dataset, so as to verify the \medit{\textbf{memorization and understanding}} of injected knowledge (the answer can be directly found in training corpora).
Then, we delve into the multi-choice question-answering task on MMedBench~\citep{qiu2024mmedlm}, to explore how LLMs \medit{\textbf{apply}} the injected knowledge in basic medicine to determine the real-world diagnosis for patients with logical reasoning.

\subsection{Preliminary Free-form QA Investigation}
\label{sec:exp_long}

\textbf{Datasets and Tasks.}
\medit{Seven subsets with 1,350 test examples from LongBench~\citep{bai2023longbench} are utilized to evaluate closed-book question-answering evaluation, where the answers can be directly found in corresponding passages.
The 14K reading-comprehension passages are used for knowledge injection via CPT/SCPT, and another 2,700 QA samples are generated for SFT/SSFT.}
Details are described in \cref{app:sec:impl_longb}.

\textbf{Evaluation Metrics.}
Here we first report the \textit{recall}~\citep{zhu2023physics32} for models' outputs against ground-truth answers to quantify the knowledge memorization degree.
In \cref{app:sec:longbench_f1} we also evaluate F1-score for a thorough comparison.

\textbf{Investigated Models.}
We mainly investigate Llama2-7B~\citep{touvron2023llama2} to compare the knowledge injection performance.

\textbf{Implementation Details.}
We train all models on LongBench passages for 3 epochs using a batch size of 128, and train for 1 epoch on synthetic SFT data to avoid overfitting.
The learning rate is 2e-5.

\textbf{Main Results.}
We first try to inject passage content into LLMs with a conventional CPT+SFT paradigm for the baseline, and use CoT instructions during testing to elicit models' memorized knowledge in their responses.
However, \cref{tab:exp_long_r} indicates such an injection approach is ineffective, as the knowledge recall is only 24.2\%.

\begin{figure}[tb]
\begin{center}
  \includegraphics[width=1.\linewidth]{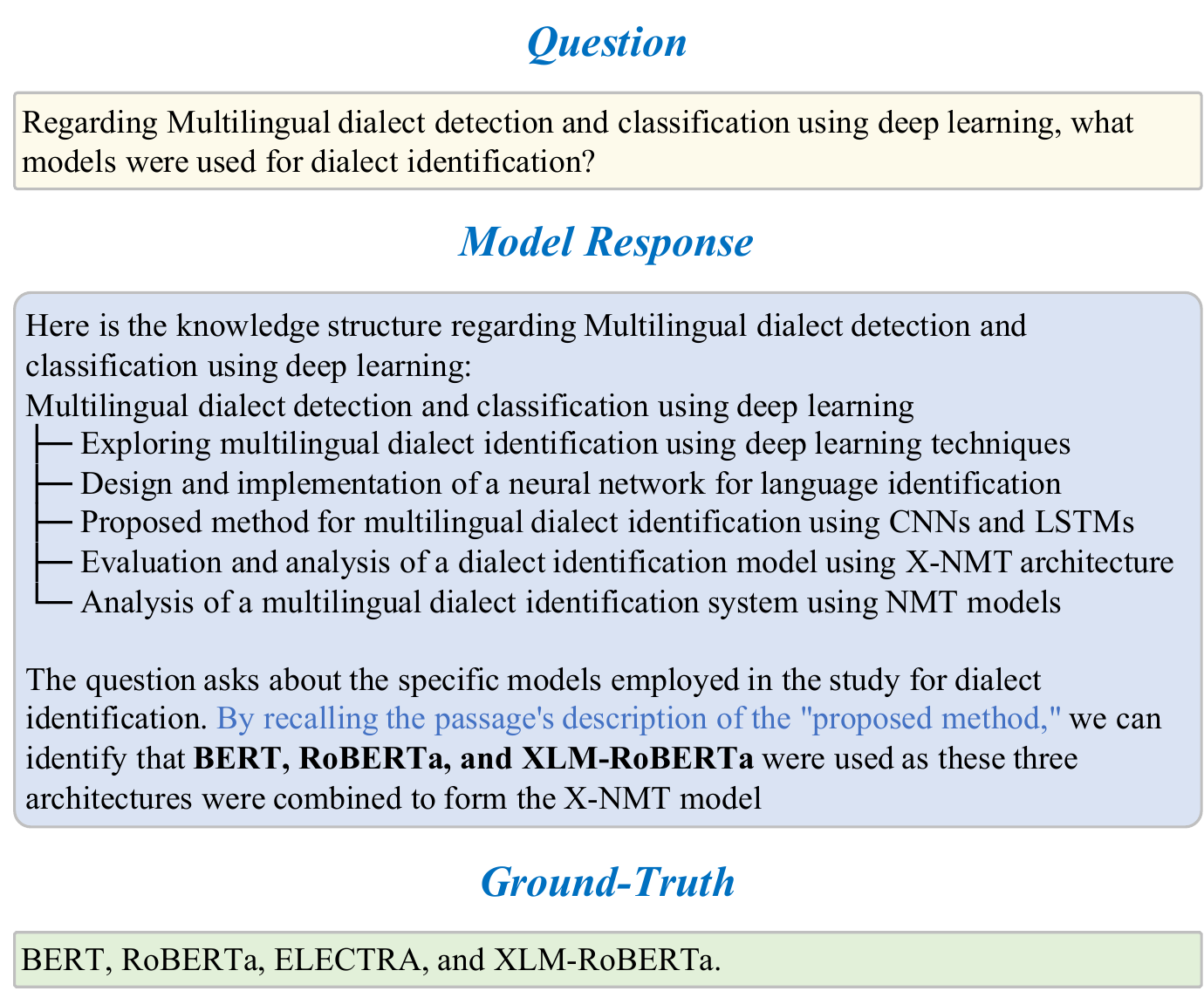}
  \caption{An example for structure-aware responses.}
  \label{fig:longb_qa_example}
\end{center}
\vskip -0.2in
\end{figure}

On the other hand, our SCPT strategy achieved a higher knowledge recall of 26.7\%. 
It implies the model has successfully associated the relevant passages with their entire knowledge structure for the given question, especially on multi-doc QAs that require more complex information retrieval and reasoning on multiple passages. Furthermore, our SSFT technique continues to improve the knowledge recall to \textbf{31.5\%}, which
indicates the vanilla SFT strategy can only regularize LLMs' response styles, while our SSFT could teach LLMs to utilize their knowledge (injected in the SCPT stage) to answer corresponding questions. \cref{fig:longb_qa_example} showcases a representative example.

In \cref{app:sec:longbench_mem}, we also validate relatively good memorization of injected knowledge structures, further emphasizing our SCPT's efficacy.

\begin{table*}[!t]
  \centering
  \caption{\medit{\textbf{Multiple-choice evaluation on MMedBench}}~\citep{qiu2024mmedlm}. We report separate accuracies across six languages, with ``Average'' denoting the mean score. ``\#Token'' denotes required data for knowledge injection. 
  }
  \label{tab:exp_mmed_full}
  \vskip -0.1in
  \setlength{\tabcolsep}{4pt}
  \small
  \begin{tabular}{l|ccccccc|c}
    \toprule
    \textbf{Model} & \textbf{English} & \textbf{Chinese} & \textbf{Japanese} & \textbf{French} & \textbf{Russian} & \textbf{Spanish} & \textbf{Average} & \textbf{\#Token}  \\ 
    \midrule
    ChatDoctor~\citep{yunxiang2023chatdoctor}     & 43.52 & 43.26 & 25.63 & 18.81 & 62.50 & 43.44 & 39.53 & -   \\
    PMC-LLaMA~\cite{wu2024pmc}      & 47.53 & 42.44 & 24.12 & 20.74 & 62.11 & 43.29 & 40.04 & -   \\
    MedAlpaca~\cite{han2023medalpaca}      & 46.74 & 44.80 & 29.64 & 21.06 & 59.38 & 45.00 & 41.11 & -   \\
    Llama2-7B~\cite{touvron2023llama2}      & 43.36 & 50.29 & 25.13 & 20.90 & 66.80 & 47.10 & 42.26 & -   \\
    InternLM2-7B~\cite{cai2024internlm2}   & 57.27 & 77.55 & 47.74 & 41.00 & 68.36 & 59.59 & 58.59 & -  \\
    \midrule
    Llama3-8B~\citep{meta2024llama3}      & 63.86 & 78.23 & 48.24 & 50.80 & 71.48 & 64.15 & 62.79~\scriptsize{\textcolor{gray}{+0.00}} & - \\
    Llama3+MMed~\cite{qiu2024mmedlm}          & \underline{66.06} & \textbf{79.25} & \textbf{61.81} & \textbf{55.63} & \underline{75.39} & {68.38} & \textbf{67.75}~\scriptsize{+4.96} & 25.5B \\
    Llama3+\mmethod~(\textbf{Ours}) & \textbf{66.77} & 77.44 & 53.27 & 51.61 & {74.61} & \underline{68.49} & {65.36}~\scriptsize{+2.57} & \textbf{76M} \\
    Llama3+\mmethod~(\textbf{Ours}) & 65.36 & \underline{79.04} & \underline{56.28} & \underline{55.47} & \textbf{80.47} & \textbf{69.80} & \underline{67.74}~\scriptsize{+4.95} & \underline{1.2B} \\
    \bottomrule
  \end{tabular} 
  \vskip -0.1in
\end{table*}

\subsection{In-depth Multi-choice QA Evaluation}
\label{sec:exp_mmed}

\textbf{Datasets and Tasks.}
\medit{We take several corpus sizes from MMedC~\citep{qiu2024mmedlm} for CPT/SCPT, with 45K QA data from MMedBench~\citep{qiu2024mmedlm}'s training set for SFT/SSFT.
Models are evaluated on six multi-choice subsets from MMedBench, where LLMs should make real-world patient diagnoses with adequate reasoning on medical knowledge.}
Detailed setup is in \cref{app:sec:impl_mmed}.

\textbf{Evaluation Metrics.}
We follow the default setting to calculate the accuracy on six language subsets and the averaged scores.
Metrics are computed by lexical exact-matching on models' responses, rather than maximum token probabilities.

\textbf{Investigated Models.}
We extend the investigation across model architectures and scales, including Llama2-7B/13B~\citep{touvron2023llama2}, InternLM2-7B~\citep{cai2024internlm2}, and Llama3-8B~\citep{dubey2024llama}.
Other popular medical LLMs~\citep{han2023medalpaca,wu2024pmc,qiu2024mmedlm} are also included for a thorough comparison.

\textbf{Implementation Details.}
Following \citet{qiu2024mmedlm}, models are first trained for 3 epochs on medical corpora with a learning rate of 2e-5, and then fine-tuned for 1 epoch to avoid overfitting.
The detailed setup is displayed in \cref{app:sec:impl_mmed}.

\begin{figure}[!t]
\begin{center}
  \includegraphics[width=1.\linewidth]{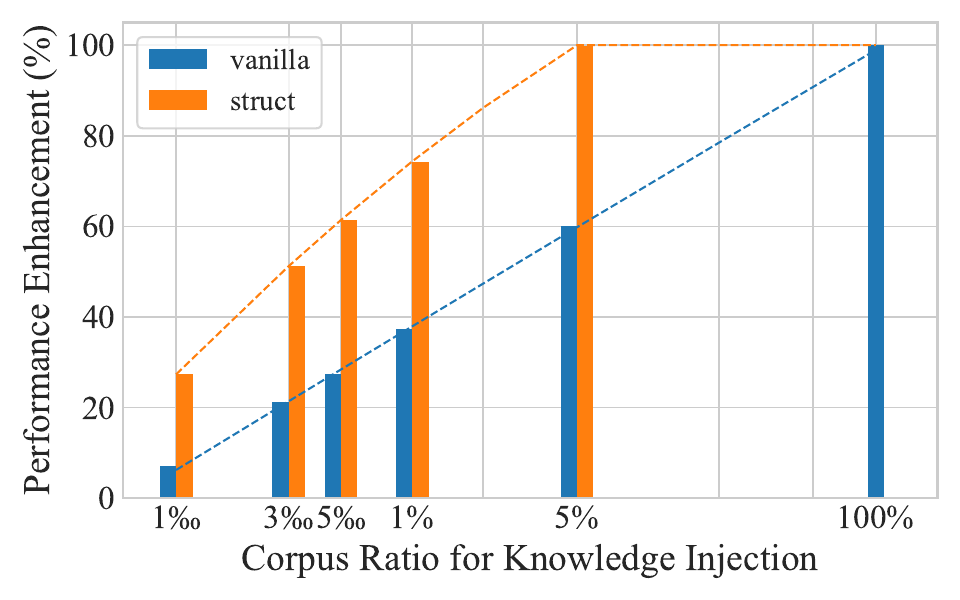}
  \vskip -0.1in
  \caption{\medit{Knowledge injection's scalability.}}
  \label{fig:mmed_scale}
\end{center}
\vskip -0.3in
\end{figure}

\textbf{Main Results.}
The results in \cref{tab:exp_mmed_full} demonstrate the promising enhancement achieved by our \mmethod~technique
largely outperforming the previous domain-specific LLMs like PMC-LLaMA~\citep{wu2024pmc} and MedAlpaca~\citep{han2023medalpaca}.
Notably, our structure-aware knowledge injection approach, using merely 76M tokens (\textbf{0.3\%}) curated from medical textbooks, achieves over \textbf{50\%} performance (2.57\% \textit{v.s.} 4.96\%) against the state-of-the-art MMedLM~\citep{qiu2024mmedlm} method, which is trained on the entire MMedC~\citep{qiu2024mmedlm} corpora of 25.5B tokens. 
\medit{After scaling up the training tokens to 1.2B (around \textbf{5\%}), our method makes nearly \textbf{100\%} improvements to the average accuracy, significantly reducing the training cost of traditional knowledge injection approaches.}

\textbf{Approach's Scalability.}
We further curate a series of training corpora sizes to investigate our method's scalability in-depth: 
30M, 76M, 132M, 250M, and 1.2B, which respectively take around 0.1\%, 0.3\%, 0.5\%, 1\%, and 5\% of 25.5B tokens.
The vanilla CPT-SFT paradigm and our SCPT-SSFT strategy are thoroughly compared across those data settings.
According to \cref{fig:mmed_scale}, our method consistently surpasses the vanilla paradigm by a large margin, emphasizing the efficacy and efficiency of domain knowledge injection.
In particular, we fit two performance-ratio scaling curves from the data points in \cref{fig:mmed_scale} as:
\setlength\abovedisplayskip{3pt}
\setlength\belowdisplayskip{3pt}
\begin{equation}
\begin{aligned}
  p_v \approx -0.04 (\log r)^2 + 13.3 \log r + 100.0 \\
  p_s \approx -1.11 (\log r)^2 + 7.63 \log r + 133.0
  \label{eq:mmed_scale}
\end{aligned}
\vspace{-2pt}
\end{equation}
\setlength\abovedisplayskip{3pt}

\noindent where $p_v$ and $p_s$ denote the relative performance enhancement (\%) for vanilla and structure-aware knowledge injection, and $r$ is the corpus ratio.

\medit{In fact, we use the scaling law derived from 0.1\%, 0.3\%, 0.5\%, and 1\% points to predict that achieving 100\% performance would require 5\% of the data corpus, which has been confirmed in \cref{tab:exp_mmed_full}.
On the other hand, it also indicates our method may lead to 133\% enhancement with a further 100\% comprehensive data utilization, further validating the effectiveness and scalability of our method.}

\begin{table}[!t]
  \centering
  \caption{\textbf{Generalization to various model architectures and sizes.} Here we use 76M training corpus.}
  \label{tab:exp_mmed_general_sim}
  \vskip -0.1in
  \small
  \begin{tabular}{lccc}
    \toprule
    \textbf{Model} & \textbf{Size} & \textbf{Adaptation} & \textbf{Accuracy}  \\ 
    \midrule
    \multirow{2}{*}{InternLM2} & \multirow{2}{*}{7B} 
    &  CPT+SFT  & 58.59  \\
    & & \textbf{SCPT+SSFT} & \textbf{63.05} \\
    \midrule
    \multirow{2}{*}{Llama2} & \multirow{2}{*}{7B} 
    &  CPT+SFT  & 42.26  \\
    & & \textbf{SCPT+SSFT} & \textbf{51.04} \\
    \midrule
    \multirow{2}{*}{Llama2} & \multirow{2}{*}{13B} 
    &  CPT+SFT  & 48.33  \\
    & & \textbf{SCPT+SSFT} & \textbf{54.50} \\
    \bottomrule
  \end{tabular} 
  \vskip -0.1in
\end{table}

\begin{table*}[!t]
  \centering
  \caption{\medit{\textbf{Ablation studies with Llama2-7B on the English subset of MMedBench.}} 
  }
  \label{tab:exp_mmed_abl}
  \vskip -0.1in
  \small
  \begin{tabular}{l|c>{\color{gray!75}}c>{\color{gray!75}}c>{\color{gray!75}}c>{\color{gray!75}}c>{\color{gray!75}}cc}
    \toprule
    \textbf{Adaptation} & \textbf{English} & \textbf{Chinese} & \textbf{Japanese} & \textbf{French} & \textbf{Russian} & \textbf{Spanish} & \textbf{Average}  \\
    \midrule
     SFT  & 44.54 & \underline{32.81} & \textbf{26.63} & 15.27 & \underline{53.91} & \underline{42.30} & \underline{35.91} \\
     CPT +  SFT  & 46.27 & 32.57 & \underline{26.13} & 17.36 & 50.00 & 40.63 & 35.49 \\
    \midrule
    \textbf{SCPT} +  SFT  & 46.50 & 32.14 & 20.10 & \underline{18.17} & \underline{53.91} & 39.97 & 35.13 \\
    \textbf{SCPT} + \textbf{SSFT}  & \textbf{49.96} & 32.63 & 22.11 & 17.52 & 51.17 & 41.28 & 35.78 \\
    \textbf{SCPT} + \textbf{SSFT*} & \underline{49.10} & \textbf{33.92} & 18.33 & \textbf{27.14} & \textbf{57.42} & \textbf{43.73} & \textbf{38.27} \\
    \midrule
    RAG  & 38.12 & 29.22 & 22.61 & 23.34 & 53.91 & 36.47 & 33.95 \\
    AdaptLLM~\citep{cheng2023adapting} & 46.79 & 33.80 & 20.60 & 14.15 & 53.12 & 42.34 & 35.03 \\
    RAFT~\citep{zhang2024raft}     & 43.60 & 32.34 & 21.11 & 14.95 & 50.39 & 42.16 & 34.09 \\
    \bottomrule
  \end{tabular} 
  \vskip -0.1in
\end{table*}

\textbf{Approach's Generalization.}
In \cref{tab:exp_mmed_general_sim}, we also validated the performance on Llama2~\citep{touvron2023llama2} and InternLM2~\citep{cai2024internlm2} model series by using 76M tokens for knowledge injection.
Our method leads to consistently significant improvements on InternLM2-7B (+4.46\%),  Llama2-7B (+8.78\%) and Llama2-13B (+6.17\%) backbone models, further demonstrating the generalizability and scalability of our \mmethod~across model architectures and sizes.
Detailed results are presented in \cref{tab:exp_mmed_general}.

\textbf{Ablation Studies.}
We perform a comprehensive ablation study on MMedBench's English subset in \cref{tab:exp_mmed_abl}. 
As suggested by \citet{qiu2024mmedlm}, we use the English textbooks~\citep{jin2020disease} (26M tokens) to compare vanilla and structure-aware CPT, paired with corresponding SFT strategies. 
In particular, ``SFT'' uses vanilla SFT with 10K QA samples from MMedBench's training split, while ``SSFT'' applies structure-aware SFT on the same questions, enhancing answers with Llama3-70B-generated knowledge explanations (\cref{sec:method_ssft}). 
``SSFT*'' further includes 8K additional structure-aware QA pairs, totaling 18K training entries. Training hyperparameters align with the main experiment.

In \cref{tab:exp_mmed_abl}, the CPT+SFT paradigm improves accuracy by 1.73\%, while SCPT with vanilla SFT achieves a higher 46.50\%. Combining SCPT with SSFT boosts performance significantly (49.96\% \textit{v.s.} 44.54\%), highlighting the importance of structured knowledge elicitation. Adding 8K extra QA pairs (``SSFT*'') further improves performance across five subsets, demonstrating a surprising cross-lingual knowledge transfer~\citep{lai2023chatgpt,qin2024multilingual}. After SSFT, LLMs effectively use knowledge injected in one language to solve problems in others, surpassing traditional SFT. Additional comparisons in \cref{app:sec:abl_sft_synth} confirm that structure-aware syntheses can enhance knowledge application better than random syntheses.

In addition, we observe that the commonly used RAG~\citep{lewis2020rag} strategy does not bring significant advantages to the MMedBench evaluation.
The main reason lies in the gap between the pre-training corpus (comprising official knowledge statements from textbooks) and evaluated QA samples (originating from practical diagnosis records).
Knowledge injection by (S)CPT and (S)SFT shows more advantages in this situation.
In-depth investigations can be found in \cref{app:sec:exp_rag}.

\medit{\textbf{Comparision with Other Methods.}
We also compare two advanced knowledge injection methods in \cref{tab:exp_mmed_abl} to further demonstrate our \mmethod's efficacy: (1) AdaptLLM~\citep{cheng2023adapting}: domain knowledge injection by appending reading comprehension QAs to each CPT chunk, and (2) RAFT~\citep{zhang2024raft}: improving LLM's robustness to domain-specific retrieval-augmented generation using noisy retrieval-augmented SFT samples.
According to the experimental results, AdaptLLM~\citep{cheng2023adapting} brings negligible improvement in the final performance (\eg, 46.79\% \textit{v.s.} 46.27\% on the English subset), indicating such a chunk-level reading comprehension augmentation during CPT cannot help LLMs capture the entire structured domain knowledge. Concurrently, RAFT~\citep{zhang2024raft} causes even worse performance, since the retrieval process introduces too many unrelated chunks and hurts LLM's QA judgments, especially when there exists a significant gap between user query and knowledge chunks in the medical diagnosis scenario.}
\section{Conclusion}
\label{sec:conc}

This work pioneers in incorporating structure-aware methodologies to enhance domain knowledge injection into large language models. 
Through a novel SCPT-SSFT paradigm, we have set a new precedent for adapting LLMs to specialized domains, and the promising and scalable results underscore the viability and potential of our method. 
We hope to inspire further research in efficient and effective domain adaptation in the LLM community, moving a step closer to models that can truly emulate human intelligence.

\noindent \textbf{Limitation.}
Our two-stage strategy introduces added computational complexity, where taxonomy extraction and data reorganization are required in the SCPT phase, and extra QA syntheses are optionally applied in the SSFT stage. 
In \cref{app:sec:exp_all}, we provide further discussion with extensive empirical experiments. Despite the additional computational overhead introduced, our method achieves greater overall benefits and can reduce the reliance on large-scale LLMs (\eg, 70B models).
We will delve into the investigations in future work.



\bibliography{custom}


\newpage
\appendix

\renewcommand\thesection{\Alph{section}}
\renewcommand\thefigure{A\arabic{figure}}    
\renewcommand\thetable{A\arabic{table}} 
\setcounter{figure}{0}  
\setcounter{table}{0}  

\section{Implementation Details}
\label{app:sec:impl}

\subsection{Detailed Setup on LongBench}
\label{app:sec:impl_longb}

\textbf{Dataset Composition.}
To focus on the investigation of knowledge injection, we choose 7 subsets from LongBench~\citep{bai2023longbench} across single- and multi-document QA tasks in English and Chinese, and the remaining synthetic or code-orientated tasks are eliminated:
\begin{itemize}[itemsep=2pt,topsep=0pt,parsep=0pt,leftmargin=20pt]
    \item \textbf{Single-Doc QA.} For single-document QA, we take three subsets from LongBench: (1) \textit{Qasper}~\citep{dasigi2021dataset}, featured by question-answering over NLP technical papers and annotated by NLP practitioners; (2) \textit{MultiFieldQA}~\citep{bai2023longbench}, manually curated from multiple data sources and annotated by Ph.D. students; and (3) \textit{MultiFieldQA-zh}, the Chinese split also provided by~\citet{bai2023longbench}, covering multiple Chinese scenarios. \textit{MultiFieldQA} contains 150 Context-Question-Answer triplets to test, and the others subsets include 200 pieces of test samples respectively. 
    \item \textbf{Multi-Doc QA.} Multi-document QA requires LLMs to extract and combine information from multiple documents to derive the answer, which is generally more challenging than single-doc QA. We take four multi-hop QA datasets: (1) \textit{HotpotQA}~\citep{yang2018hotpotqa}, containing 2-hop questions written by native speakers given two related paragraphs; (2) \textit{2WikiMultihopQA}~\citep{ho2020constructing}, involving up to 5-hop questions synthesized through manually designed templates on Wikipedia passages; (3) \textit{MuSiQue}~\citep{trivedi2022musique}, carefully composed with up to 4-hop reasoning on an increased number of supporting and distracting context evidence; and (4) \textit{Dureader}~\citep{he2017dureader}, developed based on Baidu Search and Baidu Zhidao and filtered by~\citet{bai2023longbench} to reduce the data noise.
    Each subset has 200 test samples.
\end{itemize}

\noindent
In Single-Doc QA, we extract knowledge structures for each single passage; in Multi-Doc QA, we identify the knowledge structure across multiple passages for each test sample. There are ultimate 1350 \textit{question-answer-passage(s)-(knowledge)structure} quadruples to evaluate knowledge injection approaches on LongBench.

\textbf{SFT Data-Synthesis.}
We query Llama3-70B to generate 2,700 QA examples and remove those with over 0.5 F1-Score similarity to test samples to prevent data leakage.
During inference, when the model can generate correct answers (corresponding to specific knowledge points) that haven't been seen during the SFT stage, we can ensure the knowledge is injected at the CPT stage and SFT only enhances the instruction-following capability. In practice, merely 13 out of 2700 (around 0.5\%) synthetic data have over 0.5 F1-Score and are thus filtered out from the SFT data.

\cref{tab:long_sft_sim_stat} statistics the semantic similarity (measured by BERTScore~\citep{zhang2020bertscore}) between generated and GT questions and answers, and the results emphasize there is no knowledge leakage in the generated SFT data (they share poor semantic similarity across questions, answers, and QAs).

\begin{table}[htp]
  \centering
  \caption{Similarity statistics on synthetic SFT data and LongBench's test samples.}
  \label{tab:long_sft_sim_stat}
  \vskip -0.1in
  \small
  \begin{tabular}{cccc}
    \toprule
    Target & Question & Answer & Question-Answer   \\
    \midrule
    BERTScore  & 0.277 & 0.106 & 0.093 \\
    \bottomrule
  \end{tabular} 
  \vskip -0.1in
\end{table}

\begin{table*}[!t]
  \centering
  \caption{Sample statistics on MMedBench's QA data.}
  \label{tab:mmed_test_num_stat}
  \vskip -0.1in
  \small
  \begin{tabular}{cccccccc}
    \toprule
    Split & English & Chinese & Japanese & French & Russian & Spanish & Total   \\
    \midrule
    Train  & 10,178 & 27,400 & 1,590 & 2,171 & 1,052 & 2,657 & 45,048 \\
    Test   & 1,273  & 3,426  &   199 &   622 &   256 & 2,742 & 8,518 \\
    \bottomrule
  \end{tabular} 
\end{table*}

\begin{table*}[!t]
  \centering
  \caption{\medit{Sample statistics on different training data settings.} 
  }
  \label{tab:mmed_train_num_stat}
  \vskip -0.1in
  \small
  \begin{tabular}{lcccccccc}
    \toprule
    Stage & Ratio & English & Chinese & Japanese & French & Russian & Spanish & Total   \\
    \midrule
    CPT  & 0.1\% &  6.9M &  8.8M &   -   &  4.1M &  4.9M &  5.4M & 30.1M \\
    CPT  & 0.3\% & 26.1M & 21.5M &   -   &  8.1M & 10.3M & 10.1M &  76.1M \\
    CPT  & 0.5\% & 35.9M & 27.2M &  4.5M & 14.0M & 24.9M & 26.1M & 132.6M \\
    CPT  & 1.0\% & 44.1M & 39.8M & 34.2M & 45.1M & 47.9M & 38.4M & 249.5M \\
    CPT  & 5.0\% & 169.4M & 227.8M & 119.8M & 232.7M & 235.5M & 226.1M & 1.2B \\
    \midrule
    SFT  & - & 18.8K & 39.1K &  1.6K &  5.3K &  5.9K &  7.5K &  78.2K \\
    \bottomrule
  \end{tabular} 
\end{table*}

\subsection{Detailed Setup on MMedBench}
\label{app:sec:impl_mmed}

\textbf{Data for Evaluation.}
The Multilingual Medical Benchmark (MMedBench)~\citep{qiu2024mmedlm} represents a comprehensive and diverse multilingual medical Question and Answering (QA) benchmark designed to evaluate models' capabilities of understanding and processing medical content. 

MMedBench's robust dataset extends across 6 languages (\ie, English, Chinese, Japanese, French, Russian, and Spanish) and 21 medical fields, which include, but are not limited to, Internal Medicine, Biochemistry, Pharmacology, Psychiatry, and many others. It provides 45,048 training pairs and 8,518 testing pairs for diverse learning and testing scenarios. The training split is specifically designed for domain-specific finetuning of large language models (LLMs), while the entire testing set allows for a precise assessment of multi-choice question-answering performance.
Statistics on six languages are displayed in \cref{tab:mmed_test_num_stat}.
Notably, the benchmark includes scenarios where questions may have multiple correct answers (\ie, in Japanese and French subsets), introducing additional complexity for the model evaluation process.

\textbf{Data for Continual Pre-Training.}
\medit{To investigate high-quality domain knowledge injection for LLMs and the scalability of injection methods, we curate a series of training corpus sizes from the 25.5B MMedC~\citep{qiu2024mmedlm} dataset, including 0.1\%, 0.3\%, 0.5\%, 1\%, and 5\%, which respectively takes 30M, 76M, 132M, 250M, and 1.2B tokens.
We sorted the document-level text content (including textbooks and other corpus from websites and wikipedia) by length in descending order and progressively included more samples to expand the training dataset at larger scales, where \cref{tab:mmed_train_num_stat} provides detailed statistics.
In particular, as MMedC does not provide English textbooks in its released data (due to copyright issues), we collect 18 English textbooks~\citep{jin2020disease} as an alternative for English medical knowledge injection, since they have a common OCR source with MMedC~\cite{qiu2024mmedlm}. These English textbooks take around 21.5M tokens.}

\begin{figure*}[htb]
\begin{center}
\vskip -0.3in
\centerline{\includegraphics[width=1.0\linewidth]{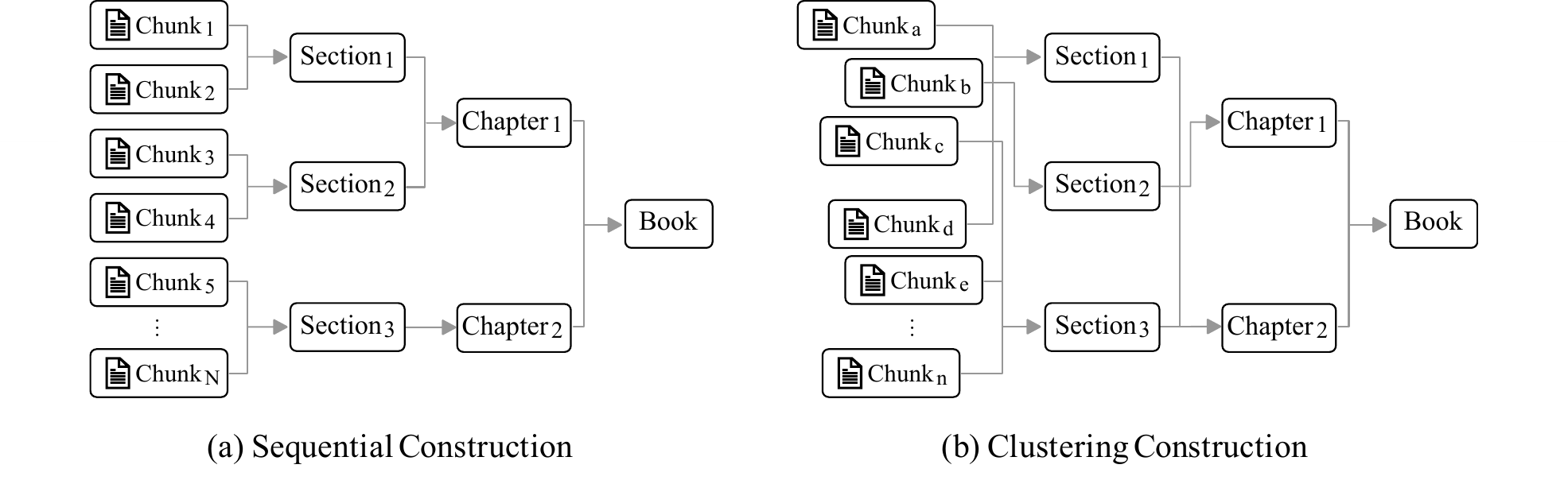}}
\caption{Knowledge structure extraction from (a) sequential chunks (\eg, from textbooks) by our specialized 7B model and (b) separated trunks (\eg, from websites) by clustering-based methods~\citep{sarthi2024raptor}. Here the terms of ``Section'', ``Chapter'', and ``book'' are just examples to help illustrate the knowledge structure.}
\label{app:fig:struct_extract}
\end{center}
\vskip -0.2in
\end{figure*}

\textbf{Knowledge Structure Extraction.}
For textbooks, we split the data to chunks (knowledge points) within 3072 tokens, and use our specifically developed 7B-size LLM to extract the structured medical knowledge system.
For non-textbook data (for instance, MMedC does not provide Japanese textbooks), a clustering-based technique~\citep{sarthi2024raptor} is adopted to recursively build knowledge structures from fragmented text segments. 
\cref{app:fig:struct_extract} presents an example to illustrate the two kinds of knowledge structure extraction processes, \medit{\cref{tab:mmed_struct_stat} displays the overall statistics on extracted knowledge structures for the final 1.2B tokens.}

\textbf{Data for Supervised Fine-Tuning.}
As introduced in \cref{sec:method_ssft}, we prompt Llama3-70B~\citep{dubey2024llama} to build the structure-aware answer explanations on top of the raw SFT samples in MMedBench's training split, and generate extra QA pairs by traversing the extracted knowledge structure from textbooks. The final quantity statistics are presented in \cref{tab:mmed_train_num_stat}.
\medit{Note that the 70B-size model is not necessary to synthesize QAs and explanations, since we have provided relevant text sources (retrieved from the training corpus) to supplement medical knowledge.
In this way, LLMs only need to perform in-context comprehension, rather than generate new QAs based on their own knowledge, which can also be achieved by smaller models like Qwen2.5-7B~\cite{yang2024qwen2}.}

\begin{table}[!t]
  \centering
  \caption{\medit{Knowledge structure on 1.2B training corpus.}}
  \label{tab:mmed_struct_stat}
  \vskip -0.1in
  \small
  \begin{tabular}{ccccc}
    \toprule
    Lang. & Book & Chapter & Section & KnowledgePoints   \\
    \midrule
    6 & 2933 & 14,411 & 23,239 & 180,793 \\
    \bottomrule
  \end{tabular} 
  \vskip -0.1in
\end{table}

\subsection{Terminology Explanation}
\label{app:define}

\textbf{Knowledge Structures.}
We extract the domain knowledge structure for each textbook, where \cref{fig:struct_isnt} presents an example, and combine the medical knowledge for six languages in MMedBench. As the (S)CPT corpus for Japanese is collected from Wikipedia rather than textbooks, we derive a single knowledge structure for Japanese medicine. 

\begin{figure*}[htb]
\begin{center}
\centerline{\includegraphics[width=0.95\linewidth]{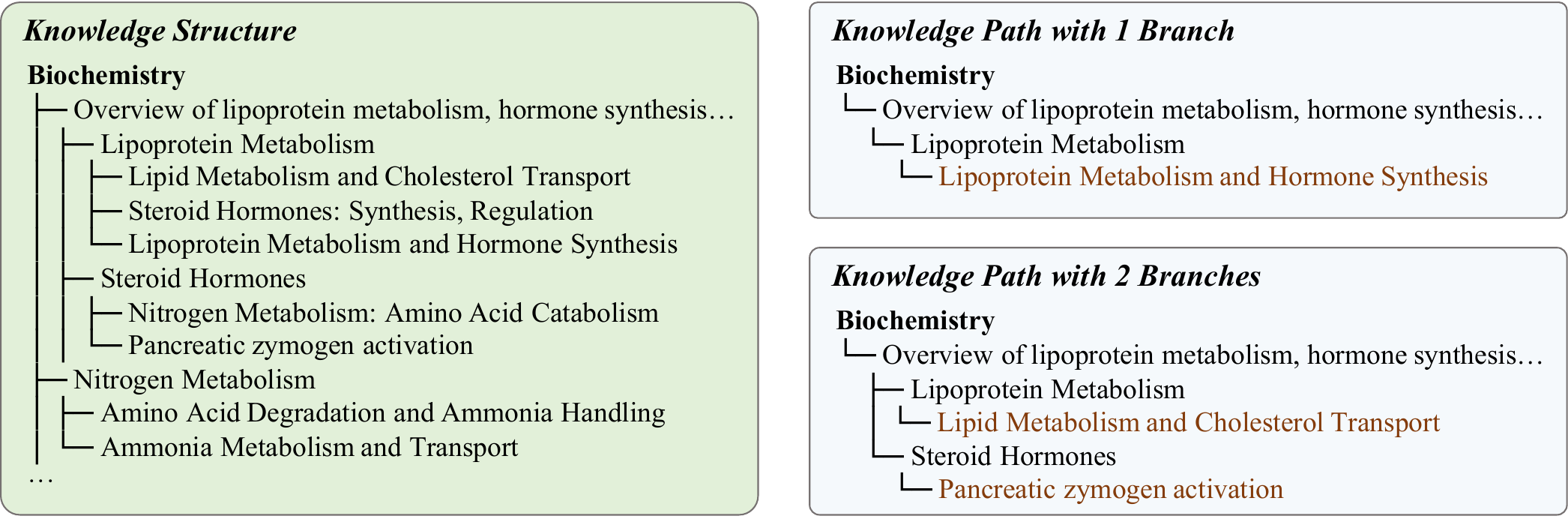}}
\caption{Definition and example of knowledge paths and branches.}
\label{fig:knowledge_path_example}
\end{center}
\vskip -0.3in
\end{figure*}

\textbf{Knowledge Paths and Branches}.
\cref{fig:knowledge_path_example} shows an example of how we define the knowledge paths and branches of the extracted knowledge structure for SSFT data synthesis.

\begin{itemize}[partopsep=2pt,itemsep=3pt,topsep=0pt,parsep=0pt,leftmargin=20pt]
    \item A \textbf{path} means a knowledge path from the domain summary (\eg, \textit{Biochemistry}) to specific knowledge points (\eg, \textit{Lipid Metabolism and Cholesterol Transport}): ``Biochemistry -- Overview of lipoprotein metabolism, hormone synthesis -- Lipoprotein Metabolism -- Lipid Metabolism and Cholesterol Transport''.
    \item A \textbf{branch} means the knowledge branch of the tree structure. If a question is related to two knowledge points (\eg, \textit{Lipid Metabolism and Cholesterol Transport} and \textit{Pancreatic zymogen activation}) at different branches of the knowledge tree, the knowledge path contains two branches, which becomes the right-bottom part of \cref{fig:knowledge_path_example}.
\end{itemize}

\subsection{Resource Requirement}
\label{app:resource}

We use 8 NVIDIA A100-80G GPUs to train all the models, and leverage 1-2 NVIDIA A100-80G GPUs for inference.
\section{Additional Experiments}
\label{app:sec:exp_all}

\subsection{Knowledge Structure Extraction}
\label{app:sec:struxgpt}

Extracting domain knowledge structure is a prerequisite for subsequent knowledge injection (including both SCPT and SSFT) for language models. In \cref{sec:method_struct}, we propose a bottom-up strategy to re-chunk the texts from domain textbooks, summarize a title for each chunk, and send the title list to a specialized 7B model to derive the knowledge structure. 
The prompt template is displayed in \cref{app:sec:struct_templ}.

In fact, we argue that due to the language diversity, a perfectly recovered table of contents of textbooks is unnecessary for domain knowledge injection. A reasonable knowledge structure is sufficient enough. 
In \cref{app:tab:exp_struxgpt}, we individually adopt few-shot GPT-3.5-Turbo~\citep{brown2020language} and LLaMA3-70B~\citep{dubey2024llama} models to extract medical knowledge structure from 18 English textbooks~\citep{jin2020disease} (with 26.1M tokens) for subsequent knowledge injection (the backbone LLM is LLaMA2-7B~\citep{touvron2023llama2}).
Although they present 3.6\%-3.7\% enhancement on MMedBench~\citep{qiu2024mmedlm}'s English test set (denoted as ``improvement''), leveraging GPT-3.5-Turbo and LLaMA3-70B is either expensive or time-consuming.
GPT-3.5-Turbo costs around 15 dollars to process 26M tokens, while LLaMA3-70B takes around 1.5 hours on 2 A100-80G GPUs, of which both limit scaling data pre-processing for structure-aware knowledge injection.

\begin{table}[!t]
  \centering
  \caption{Comparison of models to extract knowledge structures on 26M English corpus.}
  \label{app:tab:exp_struxgpt}
  \vskip -0.1in
  \renewcommand\arraystretch{1.2}
  \small
  \begin{tabular}{cccc}
    \toprule
    Model & Improvement & Time Cost & Extra Fee \\
    \midrule
     GPT-3.5-Turbo  & +3.58 & -    & 15\$ \\
     LLaMA3-70B     & +3.72 & 1.5h & - \\
     Ours-7B        & +3.69 & \textbf{0.2h} & - \\
    \bottomrule
  \end{tabular} 
  \vskip -0.1in
\end{table}

\begin{table*}[htb]
  \centering
  \caption{Structure-aware knowledge injection to Llama2~\citep{touvron2023llama2} model series.}
  \label{tab:exp_mmed_general}
  \vskip -0.1in
  \small
  \begin{tabular}{c|ccccccc}
    \toprule
    Model & English & Chinese & Japanese & French & Russian & Spanish & \textbf{Average}  \\ 
    \midrule
    InternLM2-7B   & 57.27 & 77.55 & 47.74 & 41.00 & 68.36 & 59.59 & 58.59  \\
    \textbf{+Ours} & \textbf{60.80} & \textbf{79.19} & \textbf{50.75} & \textbf{45.34} & \textbf{75.39} & \textbf{66.85} & \textbf{63.05} \\
    \midrule
    Llama2-7B      & 43.36 & 50.29 & 25.13 & 20.90 & 66.80 & 47.10 & 42.26  \\
    \textbf{+Ours} & \textbf{49.41} & \textbf{65.15} & \textbf{36.68} & \textbf{35.21} & \textbf{69.14} & \textbf{50.62} & \textbf{51.04}  \\
    \midrule
    Llama2-13B     & 51.37 & 57.97 & 32.66 & 25.08 & 69.92 & 52.99 & 48.33 \\
    \textbf{+Ours} & \textbf{53.02} & \textbf{68.30} & \textbf{37.78} & \textbf{41.71} & \textbf{70.70} & \textbf{55.51} & \textbf{54.50}  \\
    \bottomrule
  \end{tabular} 
  \vskip -0.1in
\end{table*}

Inspired by \citet{liu2024enhancing}, we distilled the knowledge structure extraction capability from giant LLMs to a LLaMA2-7B model via SFT.
In particular, we instruct LLaMA3-70B to generate 22K training examples (pairs of raw knowledge points and extracted knowledge structures) from Wikipedia, and train a LLaMA2-7B model at a batch size of 128 and a learning rate of 2e-5 for 1 epoch.
After utilizing the specialized 7B model to identify the knowledge structure in medical textbooks, as shown in \cref{app:tab:exp_struxgpt}, the results translate to comparable performance on structure-aware knowledge injection.
Meanwhile, the inference cost significantly decreases to 0.3 hours, which is more scalable to handle a larger domain corpus.

\begin{table}[!t]
  \centering
  \caption{\medit{Model evaluation on general benchmarks.}}
  \label{app:tab:train_overfit}
  \vskip -0.1in
  \renewcommand\arraystretch{1.2}
  \setlength{\tabcolsep}{6pt}
  \scriptsize
  \begin{tabular}{cccccc}
    \toprule
    Injection & MMLU & C-Eval & TruthfulQA & WinoGrande & ARC\_c \\
    \midrule
    Before & 0.46 & 0.35 & 0.49 & 0.52 & 0.51 \\
    After  & 0.43 & 0.35 & 0.43 & 0.51 & 0.56 \\
    \bottomrule
  \end{tabular} 
  \vskip -0.1in
\end{table}

\subsection{Evaluation of Approach's Generalization}
\label{app:sec:method_general}

In addition to the Llama3 models in previous experiments, we also investigate the generalization ability of our SCPT+SSFT paradigm on Llama2~\citep{touvron2023llama2} and InternLM2~\citep{cai2024internlm2} model series. As shown in \cref{tab:exp_mmed_general}, our method leads to consistently significant improvements on InternLM2-7B (+4.46\%),  Llama2-7B (+8.78\%) and Llama2-13B (+6.17\%) backbone models.
The results further demonstrate the generalizability and scalability of our \mmethod~strategy across model architectures and sizes.

\subsection{Investigation on Model Overfitting}
\label{app:sec:method_overfit}

\medit{Here we provide a supplementary evaluation on common benchmarks in \cref{app:tab:train_overfit}, which indicates the current methodology of structure-aware knowledge injection does not significantly hurt LLM's general capabilities, and even brings slight enhancement to the ARC\_c benchmark (maybe the knowledge structure enhances the reasoning ability).
As previous research states, the issue of overfitting can be further mitigated by incorporating common QA examples, and we leave this in our future work.}

\subsection{Comparison on Training Costs}
\label{app:sec:train_cost_cmp}

\begin{table}[!t]
  \centering
  \caption{Training costs for knowledge injection.}
  \label{app:tab:train_cost_cmp}
  \vskip -0.1in
  \renewcommand\arraystretch{1.2}
  \setlength{\tabcolsep}{5pt}
  \small
  \begin{tabular}{ccccc}
    \toprule
    Paradigm & Corpus & Improve. & Preprocess & Total \\
    \midrule
     CPT+SFT   & 100\%  & 100\% &   -    & $>$30d \\
     SCPT+SSFT & 0.3\%  &  50\% &  0.6h  & 4.5h \\
     SCPT+SSFT &   5\%  & 100\% &  9.7h   &  \textbf{3d} \\
    \bottomrule
  \end{tabular} 
  \vskip -0.1in
\end{table}

\begin{table*}[!t]
  \centering
  \caption{Ablation studies of SCPT on MMedBench subsets. The base model is Llama2-7B.}
  \label{tab:abl_scpt}
  \vskip -0.1in
  \renewcommand\arraystretch{1.2}
  \small
  \begin{tabular}{c|c>{\color{gray!75}}c>{\color{gray!75}}c>{\color{gray!75}}c>{\color{gray!75}}c>{\color{gray!75}}cc}
    \toprule
    Adaptation & English & Chinese & Japanese & French & Russian & Spanish & \textbf{Average}  \\
    \midrule
    CPT+SFT            & 46.27 & 32.57 & 26.13 & 17.36 & 50.00 & 40.63 & 35.49 \\
    \midrule
    Ours-FixTmpl1      & 48.27 & 32.86 & 20.61 & 23.70 & 56.17 & 42.56 & 37.36 \\
    Ours-FixTmpl2      & 48.10 & 32.99 & 21.23 & 23.97 & 55.97 & 43.10 & 37.56 \\
    \midrule
    Ours-RemoveL1      & 47.90 & 32.90 & 20.11 & 24.63 & 57.10 & 43.43 & 37.68 \\
    Ours-RemoveL2      & 48.05 & 33.63 & 21.62 & 24.11 & 57.49 & 43.13 & \underline{38.00}\\
    Ours-RemoveL3      & 48.47 & 33.14 & 20.15 & 23.33 & 56.86 & 43.65 & 37.60 \\
    \midrule
    Ours-NTPLoss       & \underline{48.99} & 33.15 & 20.57 & 25.31 & 56.78 & 42.94 & 37.96 \\
    \textbf{Ours-Full} & \textbf{49.10} & 33.92 & 18.33 & 27.14 & 57.42 & 43.73 & \textbf{38.27} \\
    \bottomrule
  \end{tabular} 
  \vskip -0.1in
\end{table*}

In \cref{app:tab:train_cost_cmp}, we quantify the total training cost on 8 A100-80G GPUs. According to \citet{qiu2024mmedlm}, the conventional CPT+SFT paradigm on 25.5B medicine corpus takes more than 30 days to derive the SOTA MMedLM model. In our SCPT+SSFT framework, although the pre-processing (\ie, knowledge structure extraction) introduces an extra 0.6 hours to process 0.3\% data (around 76M tokens), the total training process only costs 4.5 hours. As suggested in \cref{fig:mmed_scale}, when 5\% training data is leveraged for knowledge injection to achieve 100\% improvement, the overall cost is limited to 3 days, much less than the CPT+SFT approach with more than a month. Those analyses further demonstrate the efficacy and efficiency of our structure-aware knowledge injection framework.

\subsection{Ablation on Structured Knowledge Injection}
\label{app:sec:abl_mmed_scpt}

During the Structure-aware Continual Pre-Training (SCPT) stage, we proposed to learn specific text chunks (knowledge points) in the condition of the mindmap inputs (knowledge structures), in order to relate the knowledge points to corresponding structure nodes. 
In this section, we conduct a series of ablation studies to investigate the design efficacy.
The vanilla CPT+SFT paradigm is adopted as the comparison baseline, where the Llama2-7B model is trained with CPT and SFT data on the English subset of MMedBench, while tested on all subsets across six languages. The hyper-parameter settings follow the main experiment in our manuscript. The empirical results are presented in \cref{tab:abl_scpt}.

First, we investigate the choice of formatting template to convert the knowledge structure to a mindmap condition. In particular, we try to fix the template to convert all knowledge mindmaps for SCPT, and randomly select two templates to repeat the experiment. According to \cref{tab:abl_scpt}, fixed SCPT templates lead to inferior performance against randomly choosing the template from the diversified 20 template pool. This is consistent with \citet{zhu2023physics}'s observation, that text rewriting can provide better knowledge augmentation for large language models.

Then, we explore the impact of the extracted knowledge structure itself. In MMedBench, a 3-layer knowledge structure (follow the \textit{chapter-section-subsection} hierarchy) is constructed for each textbook, and we respectively remove the 1st (chapter), 2nd (section), and 3rd (subsection) layer of the hierarchy during knowledge injection. As \cref{tab:abl_scpt} shows, removing the top layer (chapter) leads to the worst performance of 47.90\%, because the remaining knowledge points cannot effectively relate to each other without the organization of the top layer. On the other hand, removing the bottom layer (subsection) performs slightly better on the English subset (because of the controlled structure-information lost), but hinders the cross-language knowledge utilization on the remaining subsets (\eg, 37.60\% on average across six languages).

Finally, we revisit the modeling choice of the mindmap-conditioning learning. Specifically, we try to turn the conditional modeling $p(\vx^k|\vs^k)$ back to complete next-token prediction $p(\vx^k, \vs^k)$ (the next-token prediction loss is computed on mindmap condition as well). According to \cref{tab:abl_scpt}, the performance is slightly inferior to our full version of SCPT strategy (\eg, 37.96\% \textit{v.s.} 38.27\% on Average). Therefore, we reserve conditional modeling for our SCPT stage.

\subsection{Ablation on SFT Data Synthesis}
\label{app:sec:abl_sft_synth}

\begin{table*}[ht]
  \centering
  \caption{Comparison of SFT data synthesis strategies on MMedBench. The backbone LLM is the same Llama2-7B model after SCPT on English textbooks.}
  \label{tab:mmed_abl_sft_synth}
  \vskip -0.1in
  \small
  \begin{tabular}{c|c>{\color{gray!75}}c>{\color{gray!75}}c>{\color{gray!75}}c>{\color{gray!75}}c>{\color{gray!75}}cc}
    \toprule
    SFT synthesis & English & Chinese & Japanese & French & Russian & Spanish & \textbf{Average}  \\
    \midrule
    -  & 46.50 & 32.14 & \textbf{20.10} & 18.17 & 53.91 & 39.97 & 35.13 \\
    SFT* & 47.13 & 32.49 & 16.58 & 16.72 & 51.95 & 42.16 & 34.51 \\
    \textbf{SSFT*} & \textbf{49.10} & \textbf{33.92} & 18.33 & \textbf{27.14} & \textbf{57.42} & \textbf{43.73} & \textbf{38.27} \\
    \bottomrule
  \end{tabular} 
\end{table*}

In \cref{sec:exp_mmed}, we compared our structure-aware knowledge injection with conventional CPT+SFT paradigm on MMedBench. On its English subset, we ablated the training components of our method, and found that the newly synthesized 8K SSFT data (by traversing the extracted knowledge structure) can inspire LLMs' cross-language capability to apply the learned structured knowledge to solve practical diagnosis problems. Here, we follow \citet{liu2024best} to randomly generate another 8K QA pairs for SFT alignment for further comparison, denoted as ``SFT*''. We randomly sample medical texts and instruct Llama3-70B~\citep{dubey2024llama} for data synthesis, without the knowledge structure provided. \cref{tab:mmed_abl_sft_synth} indicates that ``SFT*'' brings slight enhancement to the English test subset, but the average accuracy drops to 34.51\% instead. The results further demonstrate our method's efficacy in the application of the injected, structured domain knowledge.

\subsection{In-depth Comparison on Retrieval-Augmented Generation}
\label{app:sec:exp_rag}
\begin{table*}[ht]
  \centering
  \caption{Ablation on the hyper-parameter settings for the RAG baseline.}
  \label{tab:abl_rag_hyp}
  \vskip -0.1in
  \small
  \begin{tabular}{c|ccc|ccc|ccc}
    \toprule
    ChunkSize   & \multicolumn{3}{c|}{256} & \multicolumn{3}{c|}{512} & \multicolumn{3}{c}{1024}  \\
    \midrule
    RetrieveNum & 10 & 5 & 3 & 5 & 3 & 2 & 3 & 2 & 1 \\
    \midrule
    Accuracy    & 35.08 & 37.67 & 38.04 & 36.42 & \textbf{38.12} & 37.99 & 35.00 & 36.89 & 38.07
 \\
    \bottomrule
  \end{tabular} 
\end{table*}

\begin{table}[ht]
  \centering
  \caption{Attempts to integrate hybrid-search and reranker models.}
  \vskip -0.1in
  \small
  \begin{tabular}{ccc}
    \toprule
    Hybrid & Reranker & Accuracy   \\
    \midrule
    $\times$ & $\times$  & \textbf{38.12} \\
    $\surd$  & $\times$  & 37.97 \\
    $\times$ & $\surd$ & 37.52 \\
    $\surd$  & $\surd$ & 37.75 \\
    \bottomrule
  \end{tabular} 
  \label{tab:abl_rag_rerank}
  \vskip -0.1in
\end{table}

In \cref{sec:exp_mmed}, we briefly compare RAG adaptation and injection-based approaches in the MMedBench dataset, and this section provides more implementation details and further investigations on the popular retrieval-augmented generation approach.

\textbf{Experimental Settings.}
On the implementation of the RAG baseline, we utilize the BAAI/bge-m3~\citep{chen2024bge} embedding model for dense retrieval, due to its state-of-the-art and multi-lingual semantic retrieval ability. For the experiments in \cref{tab:exp_mmed_abl}, we take the same 26M English CPT data as the knowledge base, re-chunk the data corpus for every 512 tokens, and retrieve top-3 related chunks as context inputs for LLM's generation process. The retrieval process is implemented using the LlamaIndex\footnote{https://www.llamaindex.ai/} framework.

\begin{figure*}[h]
\begin{center}
\centerline{\includegraphics[width=1.0\linewidth]{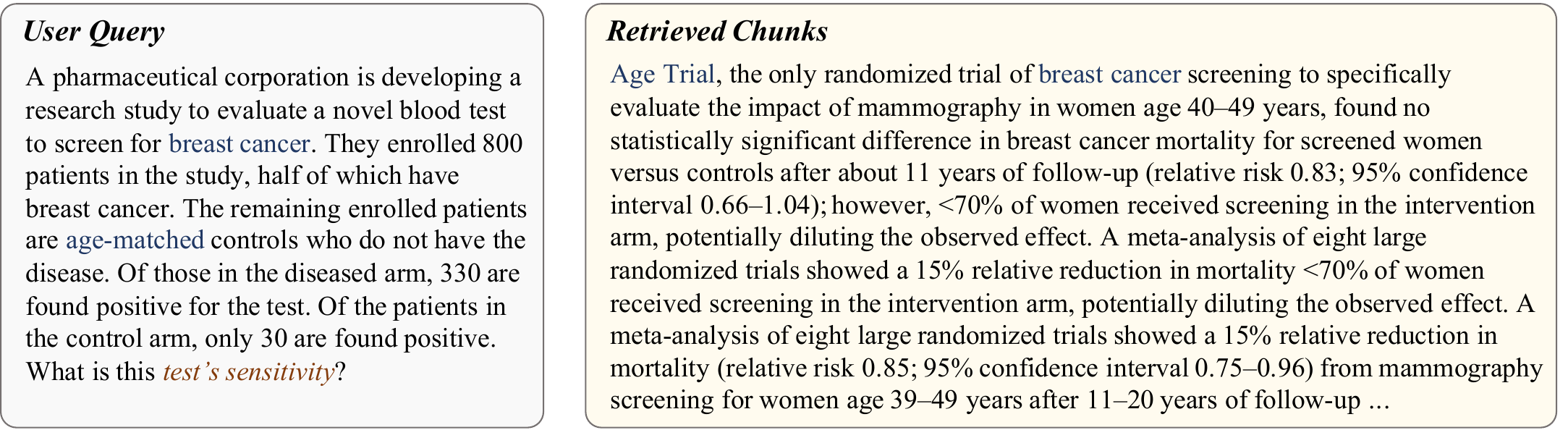}}
\caption{An example of retrieved document/chunk based on a given query.}
\label{fig:rag_example}
\end{center}
\vskip -0.3in
\end{figure*}

\textbf{Additional Experiments.}
We also conduct a variety of experiments to evaluate the hyper-parameters for the RAG baseline. As shown in \cref{tab:abl_rag_hyp}, changing the chunk size and retrieved chunk number cannot bring any significant benefits. The core reason lies in the gap between user query and retrieved chunks. In particular, user queries contain many descriptive and quantitative sentences and numbers (such as the example in \cref{fig:rag_example}, ``They enrolled 800 patients in the study, half of which have breast cancer''.), and may even talk about an entirely new thing that has not been recorded in the knowledge base.

Furthermore, we also try to use the hybrid (dense+sparse) search strategy and larger rerank model (BAAI/bge-reranker-v2-m3~\citep{chen2024bge}) to enhance the retrieval quality. However according to the results in \cref{tab:abl_rag_rerank}, the semantic gap between user queries and retrieved chunks still exists. Introducing the hybrid search and rerank model even gets worse performance (\eg, the keyword \textit{age} may be considered a key factor for hybrid search, but it cannot help to derive the answer of test sensitivity).

\textbf{Conclusion.}
RAG may assist in some knowledge-intensive tasks for information-seeking, but will encounter problems when there exists a significant semantic gap between user query and retrieved documents. MMedBench is a typical scenario, where LLMs are asked to derive medical diagnoses with proper reasoning according to the descriptions of patients or medical examinations. In this case, the retrieval process introduces too many unrelated chunks and hurts LLM's QA judgments. \cref{fig:rag_example} provides an example where the retrieved chunks are actually unrelated to the complicated user query (the user asks about the analysis of a given research study, but the retrieved documents contain several keywords, \eg, \textit{age}, while having nothing to do with the \textit{blood test study}.)

\subsection{Knowledge Memorization on LongBench}
\label{app:sec:longbench_mem}

Here, we also use lexical ROUGE-L~\citep{lin2004rouge} and semantic BERTScore~\citep{zhang2020bertscore} to quantify the memorization of injected knowledge structures, by comparing the mindmap in models' responses (as \cref{fig:longb_qa_example} displays) with ground-truth answers.
The results in \cref{tab:exp_long_mindmap} indicate a relatively good memorization of the injected knowledge mindmap, emphasizing the efficacy of our SCPT strategy.

\begin{table}[htb]
  \centering
  \caption{MindMap Recall}
  \vskip -0.1in
  \small
  \begin{tabular}{cc}
    \toprule
    F1-Score & BERTScore   \\
    \midrule
    0.61  &  0.87 \\
    \bottomrule
  \end{tabular} 
  \label{tab:exp_long_mindmap}
\vskip -0.1in
\end{table}

\subsection{F1-Score Evaluation on LongBench}
\label{app:sec:longbench_f1}

\begin{table*}[htb]
  \centering
  \caption{F1 Score evaluation of Closed-Book QA (CBQA) tasks on the LongBench~\citep{bai2023longbench} dataset. The best results are marked in \textbf{bold}, and the secondary results are marked with \underline{underlines}. The backbone model is Llama2-7B~\citep{touvron2023llama2}.}
  \label{tab:exp_long_f1}
  \vskip -0.1in
  \renewcommand\arraystretch{1.2}
  \small
  \begin{tabular}{c|c|cccccccc}
    \toprule
    \multirow{2}[1]{*}{Task} & \multirow{2}[1]{*}{Adaptation} & \multicolumn{3}{c}{SingleDoc-QA} & \multicolumn{4}{c}{MultiDoc-QA} & \multirow{2}[1]{*}{Average} \\
    \cmidrule(lr){3-5} \cmidrule(lr){6-9}
    & & Qasper & MFQA & MFQAzh & HpQA & 2Wiki & Musiq & Duzh   \\
    \midrule
    \multirow{3}[2]{*}{CBQA}
    &  CPT+SFT  & \underline{16.8} & \underline{23.1} & 13.2 & \underline{21.3} & 19.1 & \underline{10.4} & 13.4 & \underline{16.8} \\
    & SCPT+SFT  & 15.2 & 21.5 & \underline{15.2} & 14.9 & \underline{19.8} &  5.6 & \underline{14.1} & 15.2 \\
    & SCPT+SSFT & \textbf{19.7} & \textbf{23.5} & \textbf{19.5} & \textbf{26.4} & \textbf{24.1} & \textbf{12.2} & \textbf{15.4} & \textbf{20.1} \\
    \bottomrule
  \end{tabular} 
\end{table*}

In \cref{sec:exp_long}, we mainly follow \citet{zhu2023physics32} to investigate the memorization and understanding of injected knowledge by calculating the knowledge recall in models' responses. 
Here we report the F1-score measure over the Closed-Book QA (CBQA) settings for a thorough comparison.
Note that here we use the vanilla question prompt to obtain concise answers, instead of the CoT prompt used in \cref{sec:exp_long} to elicit models’ memorized knowledge. The evaluated models are the same as \cref{sec:exp_long}.

In \cref{tab:exp_long_f1}, we report the Closed-Book QA (CBQA) baseline by traditional CPT+SFT to inject passage contents into model parameters, and supplement the experiment of our SCPT+SSFT technique for comparison.
According to the results shown in \cref{tab:exp_long_f1}, 
our SCPT+SSFT approach successfully boosts the closed-book QA performance to 20.1\% on average.
The results are consistent with \cref{sec:exp_long}, which jointly demonstrate the effectiveness of structure-aware knowledge injection for large language models.

\section{Detailed Related Work}
\label{app:sec:related_work}

\indent \textbf{Domain Adaptation for LLMs}.
While pre-trained LLMs possess promising capabilities, their performance is often hampered by the scope and recency of their training data, which particularly affects smaller models in downstream applications~\citep{zhao2023survey,wang2023survey}. 
Continual Pre-Training (CPT) addresses this by perpetually updating a pre-trained model with domain-specific content~\citep{sun2020ernie,xu2023kilm}.
, with parameter-efficient tuning methods devised to curtail training costs~\citep{hu2021lora,liu2024dora}. 
To keep pace with the latest information, models can be fine-tuned with supervised instruction-response pairs (SFT), thus staying current with the advancing knowledge landscape~\citep{mecklenburg2024injecting,qiu2024mmedlm}. 
Existing literature confirms that combining CPT and SFT is effective for LLMs to remain precise and up-to-date in dynamic fields like 
medicine~\citep{wang2023huatuo,qiu2024mmedlm} 
and coding~\citep{roziere2023codellama,guo2024deepseekcoder}. 
Our study builds upon this CPT-SFT framework, innovating with SCPT-SSFT strategies to efficiently and effectively infuse domain knowledge with the inherent structure hierarchy.

\textbf{Conditional Language Modeling}.
The idea of continual pre-training language models on domain corpus in the condition of the knowledge structure is mainly inspired by CTRL~\citep{keskar2019ctrl}.
\citet{keskar2019ctrl} demonstrates the effectiveness of steering text generation through control codes (one or two words) that signify the desired genre, style, or task.
In the era of LLM, system prompt plays a similar role in controlling models' responses to adapt to different needs and functionalities, such as role-playing, language style transfer, task setting, and behavior setting~\citep{brown2020language,wang2023rolellm,bai2023qwen}. 
Our SCPT approach extends the control codes or system prompts to domain-specific knowledge structures, so as to guide the learning process and tailor the model's output more closely to specialized fields.

\textbf{Structure-aware Knowledge Aggregation}.
Knowledge structure has been widely explored in the recent LLM community. 
In conventional paradigms, researchers extract entity-relation-entity triplets from texts to construct knowledge graphs~\citep{pan2024unifying}, to enhance LLMs's factual knowledge and logical reasoning by feature aggregation~\citep{liu2020k,zhang2022dkplm}, prompt engineering~\citep{wen2023mindmap,wang2023boosting}, 
information searching~\citep{logan2019barack},  
training data synthesis~\citep{tang2024mathscale}, etc. 
In these cases, each node corresponds to either a specific entity or an abstract concept, lacking the capability to present an informative and self-contained \textit{knowledge point}.
Some works have recently related a piece of descriptive text to a knowledge point, and constructed the knowledge structure for LLMs' retrieval-augmented generation~\citep{sarthi2024raptor,dong2024multi}, where the top-to-down retrieval provides precise information-seeking paths along the knowledge structure.
In this paper, we extend the structure-aware knowledge aggregation to LLMs' training phase, injecting the whole domain knowledge structure into LLMs' by linking training samples to corresponding knowledge points and reasoning paths.

\textbf{Data Augmentation and Synthesis}.
Due to the lack of high-quality datasets, data augmentation has emerged as a promising solution to mimic real-world patterns~\citep{liu2024best}. 
Traditional methods aim to artificially expand the training dataset size~\citep{xu2023wizardlm,mukherjee2023orca} or generate entirely new samples that could help models learn better or adapt to specific tasks~\citep{tang2024mathscale}. 
Yet, they often overlook the structured nature of domain knowledge, and the aimlessly generated samples may also lack diversity~\citep{ovadia2023fine,mecklenburg2024injecting}, leading to potentially suboptimal training outcomes when applied for domain adaptations~\citep{mecklenburg2024injecting,tang2024mathscale}.
By contrast, our SSFT design is an innovative departure to address the challenge of retaining and utilizing the structured knowledge inherent in domain-specific content.
\section{Prompt Template for Knowledge Structure Extraction}
\label{app:sec:struct_templ}

\begin{figure*}[htb]
\begin{center}
\centerline{\includegraphics[width=1.0\linewidth]{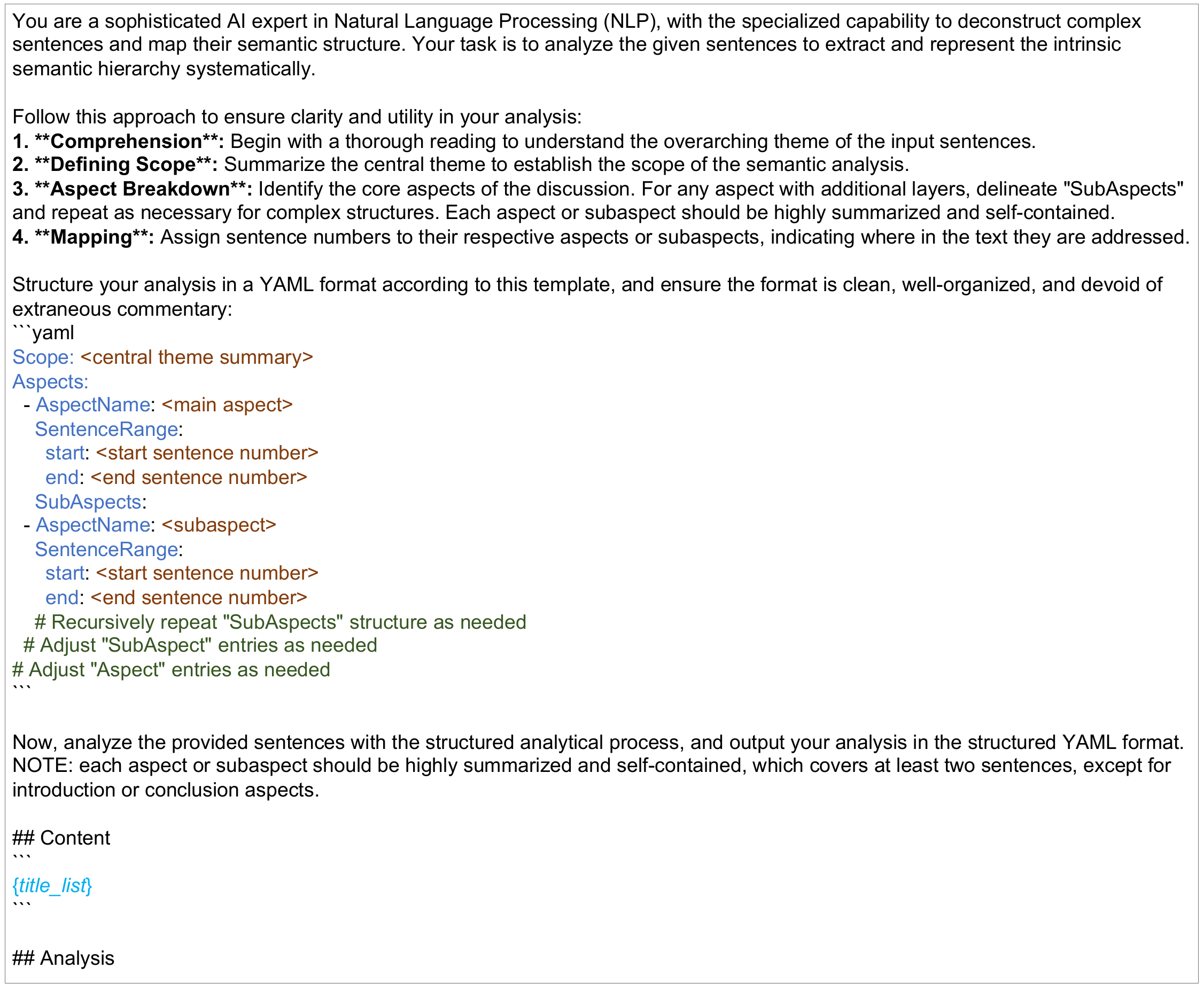}}
\caption{Prompt template for knowledge structure identification.}
\label{tab:struct_prompt_templ}
\end{center}
\vskip -0.2in
\end{figure*}

\cref{tab:struct_prompt_templ} displays the prompt template to query our specialized 7B model to extract knowledge structure on given knowledge points, which introduces the task definition, detailed instruction, and output formats to illustrate the process.

\begin{figure*}[htb]
\begin{center}
\centerline{\includegraphics[width=0.98\linewidth]{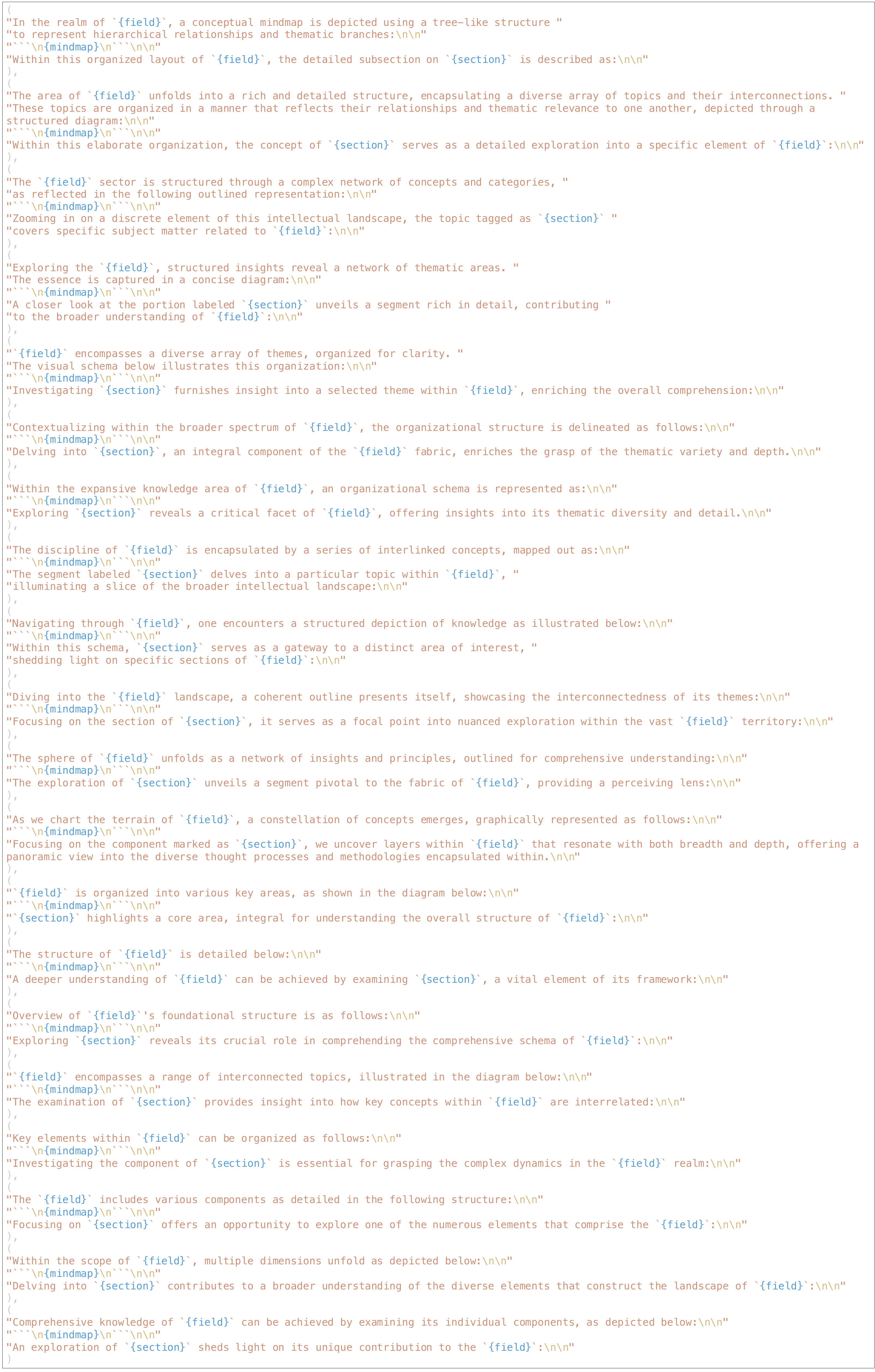}}
\caption{Full template pool for mindmap conversion with 20 diversified templates.}
\label{fig:mindmap_templates}
\end{center}
\vskip -0.2in
\end{figure*}


\end{document}